\documentclass[twoside,11pt]{article}
\usepackage{jmlr2e}

\usepackage[utf8]{inputenc} 
\usepackage[T1]{fontenc}    
\usepackage{url}            
\usepackage{booktabs}       
\usepackage{amsfonts}       
\usepackage{amsmath}
\usepackage{amssymb}
\usepackage{nicefrac}       
\usepackage{microtype}      
\usepackage{subfigure}
\usepackage{mathtools}
\usepackage{bm}
\usepackage{here}
\usepackage{color}

\usepackage{nicefrac}
\usepackage{listings}
\usepackage{docmute}
\usepackage[acronym, nomain]{glossaries}
\usepackage{indentfirst}
\usepackage{algorithm}
\usepackage{algorithmic}
\usepackage{wrapfig}

\usepackage{microtype}
\usepackage{subfigure}
\usepackage{booktabs} 
\usepackage{amsmath}
\usepackage{amssymb}
\usepackage{amsfonts}
\usepackage{mathtools}
\usepackage{bm}
\usepackage{here}

\usepackage{natbib}

\usepackage{wrapfig}
\usepackage{algorithm}
\usepackage{algorithmic}
\usepackage{mathtools}
\usepackage{verbatim}
\usepackage{appendix}

\newtheorem{rem}{Remark}
\newtheorem{lem}{Lemma}
\newtheorem{thm}{Theorem}
\newtheorem{prop}{Proposition}

\newtheorem{assump}{Assumption}

\newcommand{\diff}[1]{{\textcolor{black}{#1}}}

\newcommand{\argmin}{\operatornamewithlimits{argmin}}

\usepackage{lastpage}
\jmlrheading{22}{2021}{1-\pageref{LastPage}}{6/20; Revised
10/21}{11/21}{20-653}{Masahiro Fujisawa and Issei Sato}
\ShortHeadings{Multilevel Monte Carlo Variational Inference}{Fujisawa and Sato}

\firstpageno{1}

\begin{document}

\title{Multilevel Monte Carlo Variational Inference}

\author{\name Masahiro Fujisawa \email fujisawa@ms.k.u-tokyo.ac.jp \\
\addr Graduate School of Frontier Sciences\\
The University of Tokyo\\
5-1-5 Kashiwanoha, Kashiwa, Chiba 277-8561, Japan \\
\addr Center for Advanced Intelligence Project\\
RIKEN\\
1-4-1 Nihonbashi, Chuo-ku, Tokyo 103-0027, Japan
\AND
\name Issei Sato \email sato@k.u-tokyo.ac.jp \\
\addr Graduate School of Information Science and Technology\\
The University of Tokyo\\
7-3-1 Hongo, Bunkyo-ku, Tokyo 113-8656, Japan
}

\editor{Erik Sudderth}

\maketitle

\begin{abstract}
We propose a variance reduction framework for variational inference using the Multilevel Monte Carlo (MLMC) method. Our framework is built on reparameterized gradient estimators and ``recycles'' parameters obtained from past update history in optimization. In addition, our framework provides a new optimization algorithm based on stochastic gradient descent (SGD) that adaptively estimates the sample size used for gradient estimation according to the ratio of the gradient variance.
We theoretically show that, with our method, the variance of the gradient estimator decreases as optimization proceeds and that a learning rate scheduler function helps improve the convergence.
We also show that, in terms of the \textit{signal-to-noise} ratio, our method can improve the quality of gradient estimation by the learning rate scheduler function without increasing the initial sample size.
Finally, we confirm that our method achieves faster convergence and reduces the variance of the gradient estimator compared with other methods through experimental comparisons with baseline methods using several benchmark datasets.
\end{abstract}


\section{Introduction}

\label{Introduction}
Variational inference (VI)~\citep{Jordan99} has been successfully used in the context of approximate Bayesian inference.
The object of VI is to seek a distribution from a variational family of distributions that best approximates an intractable posterior distribution~\citep{Miller17}.
Unfortunately, the objective function of VI itself is often intractable and cannot be optimized in a closed form in modern complex models such as Bayesian neural networks.
In this case, we often optimize the objective function based on the stochastic gradient estimated on the Monte Carlo samples from the variational distribution, which is called Monte Carlo variational inference (MCVI) or black box variational inference~\citep{Ranganath14}.
However, the stochastic gradient obtained with Monte Carlo approximation causes a slow convergence owing to the high variance.
Therefore, the variance of the stochastic gradient estimator should be controlled carefully to make MCVI useful.
 
There are two standard MCVI gradient estimators: the score function gradient estimator~\citep{Paisley12, Ranganath14} and the reparameterized gradient estimator~\citep{Titsias14,Rezende14,Kingma14}.
The score function gradient estimator is a general tool for MCVI, which can be applied to both discrete and continuous random variables; however, it often has high variance, and the training becomes difficult~\citep{Buchholz18a}.
In contrast, the reparameterized gradient estimator often has a lower variance for continuous random variables, although it is restricted to models where the variational family can be reparametrized via a differentiable mapping.
Recently, there has been extensive research on making the reparameterized gradient estimator a more general tool, including a unified view of these two gradient estimators provided by \citet{Ruiz2016b}.
For example, \citet{tokui16} and \citet{Jang17} proposed a reparameterization trick for discrete or categorical variables.
Furthermore, \citet{tucker17} proposed a low-variance reparameterized gradient estimation method for discrete latent variables.
\cite{Will18} presented a general method for unbiased gradient estimation of black-box functions
and applied it to discrete variational inference and reinforcement learning.
Moreover, the theoretical properties of the reparameterized gradient have been actively analyzed recently~\citep{Xu19, Domke19}.
Owing to these studies, the use of the reparameterized gradient has become a more practical way to reduce the variance of gradient estimators.
 
To reduce the variance of the score/reparameterized gradient estimators, we often use control variates~\citep{Glasserman03,Miller17,Geffner20} or importance sampling~\citep{Ruiz16, Burda16, Sakaya17,Li18}.
However, their use requires the construction of appropriate auxiliary functions to achieve effective variance reduction, and thus their performance depends on such a heuristic selection.
Recently, \citet{Buchholz18a} have proposed a variance reduction scheme using the randomized quasi-Monte Carlo (RQMC) method for MCVI, which does not require the design of an auxiliary function and improves the order of the conventional gradient estimator variance, $\mathcal{O}(N^{-1})$, to $\mathcal{O}(N^{- 2})$ in the best case.
Although the directions proposed in this study are extremely important for making MCVI more practical, the advantage of the RQMC method is less than theoretically expected when the integration is not sufficiently smooth and/or its dimension is high~\citep{Morokoff94}.
Furthermore, the performance of QMC-based methods sometimes becomes worse than that of MC methods owing to a potentially unfavorable interaction between the underlying deterministic points and the function to be estimated~\citep{Lemi09}.
Therefore, it is necessary to establish a \emph{sampling-based} variance reduction method that achieves a stable performance against the problematic factors described above.
 
In this paper, we propose a novel \emph{sampling-based} variance reduction framework for MCVI that ``recycles'' past gradients and parameters based on the Multilevel Monte Carlo (MLMC) method.
Our method is naturally derived from the reparameterized gradient estimator and is also independent of the design of auxiliary functions for variance reduction.
Furthermore, our method theoretically guarantees convergence acceleration and the quality of the gradient estimator.
In addition, several experiments confirm that our method converges faster, reduces variance better, and sometimes achieves better prediction performance than the baseline methods.
Since MLMC is a pure MC sampling method, our method can be combined with various variance reduction methods that have been proposed~\citep{Ranganath14,Ruiz16, Burda16, Sakaya17, Miller17,Li18,Tucker19}, but it is unclear whether this is possible with RQMC-based methods~\citep{Buchholz18a}.
We note that our framework can be easily implemented in modern inference libraries such as Pytorch~\citep{pytorch}.
Our contributions are as follows.
\begin{figure}[t]
 \centering
 \includegraphics[scale=0.47]{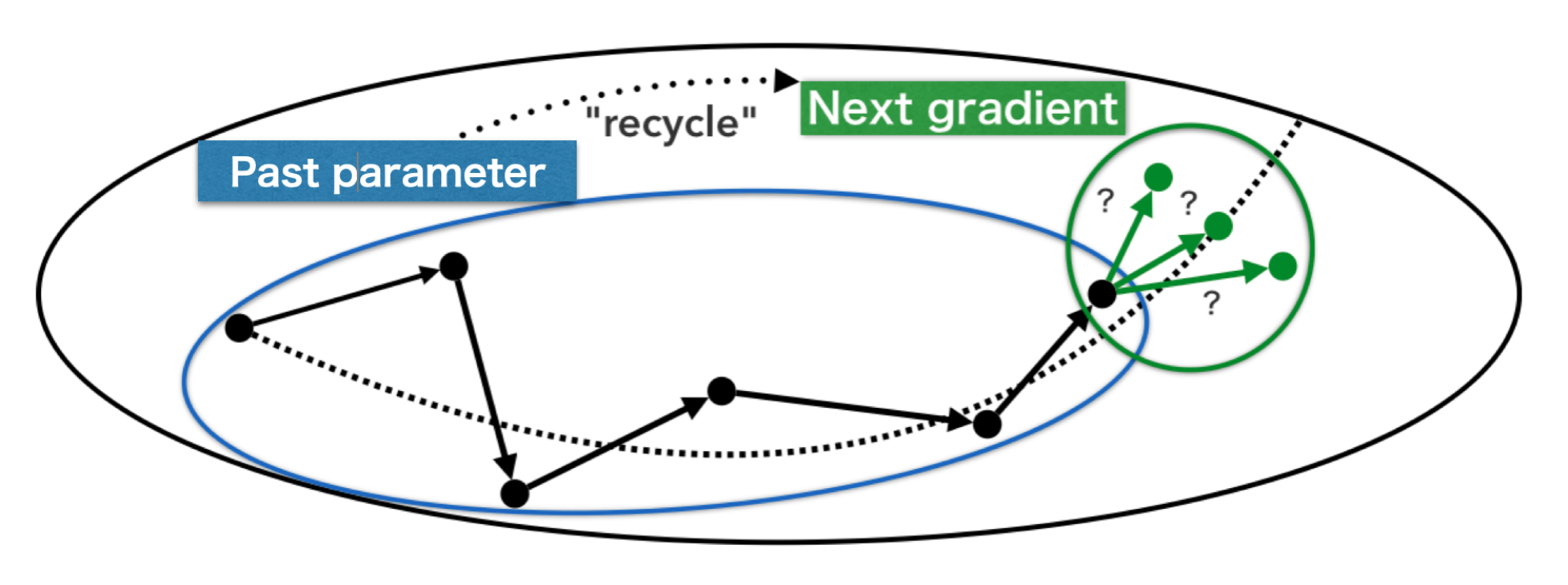}
 \caption{Concept of our method. In the context of traditional MCVI, the gradient from the current point (the rightmost point in the blue circle) is estimated with variance reduction techniques, and the parameters are updated (moved to the next green point in the green circle).
 On the other hand, our method uses the history of the parameters (all points in blue circles) to estimate the gradient with low variance and update the parameters.}
 \label{fig:recycle}
\end{figure}
\begin{itemize}
   \item We investigate the idea of using the MLMC method for MCVI on reparameterized gradient estimation.
   \item On the basis of the above idea, we develop an algorithm that provides a low-variance gradient estimator with a small number of samples by ``recycling'' the parameters in the past (see Figure \ref{fig:recycle}).
   \item Our convergence analysis shows that, in our method, the learning rate scheduler function helps accelerate the optimization compared with the baseline method.
   \item We also show that our method can improve the quality of gradient estimation by the learning rate scheduler function in terms of the \textit{signal-to-noise} ratio (SNR).
\end{itemize}
 
The rest of this paper is organized as follows.
We overview the background and related studies in Section~\ref{background}.
In Sections~\ref{mlmcvi-sec} and \ref{sec:theoretical_analysis}, we introduce our framework and discuss its theoretical properties.
Finally, we give experimental results and conclusions in Sections~\ref{exp-sec} and \ref{conclusion}.
\section{Background}
\label{background}
Here, we briefly review VI, the reparameterized gradient, and MCVI in Sections~\ref{VI}, \ref{repara}, and \ref{MCVI}.
Furthermore, in Section~\ref{related_work}, we introduce several related studies of variance reduction in the MCVI context.

\subsection{Variational Inference}
\label{VI}
The aim of Bayesian inference is to estimate the posterior distribution of latent variables ${\bf z}$ given the observation ${\bf x}$: $p({\bf z}|{\bf x})$.
The exact computation of $p({\bf z}|{\bf x})$ amounts to a sum or integration over all ${\bf z}$.
However, modern Bayesian statistics rely on a complex (e.g., nonconjugate) model for which the posterior is difficult to compute.
Furthermore, the entire inference procedure can be large-scale, and thus performing exact inference is typically intractable.
To overcome these problems, we need efficient algorithms for approximating the posterior.
 
With VI, we can construct an approximation by minimizing the Kullback--Leibler (KL) divergence between the variational distribution $q({\bf z}|\lambda)$ and $p({\bf z}|{\bf x})$, where $\lambda \in \mathbb{R}^{d}$ is a single vector of all free parameters and $d$ is the dimension of the parameter space.
{This is equivalent to} maximizing the evidence lower bound (ELBO)~\citep{Jordan99,zhang17,Blei17b}:
\begin{align}
  \label{eq:ELBO}
  \mathcal{L}(\lambda) = \mathbb{E}_{q({\bf z}|\lambda)}[\log p({\bf x,z}) - \log q({\bf z}|\lambda)],
\end{align}
{or minimizing the variational free energy~\citep{nakajima15a}:
\begin{align}
\label{eq:free_energy}
  \mathcal{F}(\lambda) = \mathbb{E}_{q({\bf z}|\lambda)}[\log q({\bf z}|\lambda) - \log p({\bf x,z})].
\end{align}
}
When VI is applied to large-scale data or a complex model, it is difficult or even sometimes impossible to directly compute the differentiation of the objective in Eqs.~\ref{eq:ELBO} {and \ref{eq:free_energy}} with respect to $\lambda$.
One way to handle this problem is to use a stochastic gradient on the basis of two major gradient estimators: the score function gradient estimator~\citep{Ranganath14} and the reparameterized gradient estimator~\citep{Titsias14,Kingma14,Rezende14}.
These gradient estimators are obtained by approximating the expectation of the gradient with independent and identically distributed (i.i.d.) samples from $q({\bf z}|\lambda)$.
{However, the score function gradient estimator tends to have a higher variance than the reparameterized gradient estimator~\citep{Miller17,Roeder17,Buchholz18a}.
That is, the reparameterized gradient estimator has attracted considerable attention as a more convenient tool to reduce the variance than the score function gradient estimator, and has been the subject of extensive research in recent years~\citep{Miller17,tucker17,Domke19,Xu19,Geffner20}.
}

\subsection{Reparameterized Gradient}
\label{repara}
The use of the reparameterized gradient is a notable approach for learning complex models or reducing the estimation variance based on the reparameterization trick~\citep{Kingma14}.
In this gradient, the variable ${\bf z}$ generated from the distribution $q({\bf z}|\lambda)$ is expressed as a deterministic transformation $\mathcal{T}(\cdot)$ of another simple distribution $p(\epsilon)$ over a noise variable $\epsilon$.
Therefore, $\bf z$ can be expressed as ${\bf z} = \mathcal{T}(\epsilon; \lambda
)$, where $\epsilon \overset{\mathrm{i.i.d.}}{\sim} p(\epsilon)$.
If we assume that $q({\bf z}|\lambda)$ is in a multivariate location-scale family,
we often use an affine transformation $\mathcal{T}(\epsilon; \lambda) = {\bf m} + {\bf v}\epsilon$~\citep{Titsias14}, where $\lambda = ({\bf m, v})$.
If we set $p(\epsilon)$ as the standard normal Gaussian $\mathcal{N}(0,I_{d})$, for example, $\mathcal{T}(\epsilon; \lambda)$ provides samples from $\mathcal{N}({\bf m, v^{\top}v})$.
By using the reparameterization trick, we can express the gradients of the ELBO {and the variational free energy} as the expectation with respect to $p(\epsilon)$ instead of $q({\bf z}|\lambda)$, {that is,}
\begin{align*}
  \mathbb{E}_{p(\epsilon)}[\nabla_{\lambda} \log p( {\bf x},\mathcal{T}(\epsilon; \lambda
   )) - \nabla_{\lambda} \log q(\mathcal{T}(\epsilon; \lambda
   )| \lambda))],
 \end{align*}
and
\begin{align}
  \label{eq:RG_free}
  \mathbb{E}_{p(\epsilon)}[\nabla_{\lambda} \log q(\mathcal{T}(\epsilon; \lambda
   )| \lambda)) - \nabla_{\lambda} \log p( {\bf x},\mathcal{T}(\epsilon; \lambda
   ))].
 \end{align}
{In the above, the distribution for calculating the expectation is fixed,} and the gradient estimator is obtained by approximating the expectation with i.i.d. random variables $\epsilon$ from $p(\epsilon)$.
{We show that this property is important for the derivation of our method in Section~\ref{sec:idea_mlmcvi}.}

\subsection{Monte Carlo Variational Inference (MCVI)}
\label{MCVI}
{From here, instead of ELBO, we focus on the variational free energy to derive our method based on gradient-descent-based optimization.}
In the general MCVI framework, 
the gradient of the {variational free energy} is represented as an expectation: $\nabla_{\lambda} \mathcal{F}(\lambda) = \mathbb{E}[ g_{\lambda}(\tilde{{\bf z}})]$ over a random variable $\tilde{{\bf z}}$, where $g_{\lambda}(\cdot)$ is a function of the gradient {in} Eq.~\ref{eq:free_energy}.
For the reparameterization estimator, Eq.~\ref{eq:RG_free} with $\tilde{{\bf z}} = \epsilon$ leads to the expression $\nabla_{\lambda}\mathcal{F}(\lambda) = \mathbb{E}_{p(\epsilon)}[g_{\lambda}(\epsilon)]$, where 
\begin{align*}
     {g_{\lambda}(\epsilon) = \nabla_{\lambda} \log q(\mathcal{T}(\epsilon; \lambda)| \lambda) - \nabla_{\lambda} \log p({\bf x},\mathcal{T}(\epsilon; \lambda)).}
 \end{align*}
To estimate this gradient stochastically, we use an \emph{unbiased} estimator calculated by averaging over i.i.d. samples $\{ \epsilon_{1},\epsilon_{2}, \ldots, \epsilon_{N} \}$, {that is,}
 \begin{align}
 \label{eq:g_lambda_estimator}
  \widehat{\nabla}_{\lambda_{t}}\mathcal{F}(\lambda_{t}) =  \hat{g}_{\lambda_{t}}(\epsilon_{1:N}) = \frac{1}{N} \sum_{n=1}^{N} g_{\lambda_{t}}(\epsilon_{n}),  
 \end{align}
where $t$ represents the optimization step.
The {variational free energy} can then be optimized on the basis of $\hat{g}_{\lambda_{t}}(\epsilon_{1:N})$ by using some form of stochastic optimization such as {stochastic gradient descent (SGD)} updates with a decreasing learning rate $\alpha_{t} = \alpha_{0}\eta_{t}$:
 \begin{align}
 \label{eq:SGD}
  \lambda_{t+1} = \lambda_{t} - \alpha_{t} \hat{g}_{\lambda_{t}}(\tilde{{\bf z}}).
 \end{align}
Here, $\alpha_{0}$ is the initial learning rate and $\eta_{t} \ (>0)$ is the value of the learning rate scheduler function, where $\eta_{0} = \eta_{-1} = 1$.
The learning rate scheduling has been empirically shown to be useful in improving inference performance~\citep{Li20} and is often used to guarantee the convergence of stochastic optimization when the gradient estimator is noisy~\citep{bottou16}.

\subsection{Other Related Studies}
\label{related_work}
Here, we introduce several related studies that have been conducted on variance reduction and SNR for MCVI.
Table~\ref{Related work} summarizes the relationships between our study and existing studies.

\paragraph{Variance Reduction for MCVI :}
Since the introduction of MCVI, VI has become a powerful tool for inference on various model architectures.
However, MCVI has the crucial problem that the convergence of the stochastic optimization scheme tends to be slow when the magnitude of the gradient estimator's variance becomes high owing to the Monte Carlo estimation.
Many techniques have been proposed for variance reduction in this context, such as using control variates~\citep{Glasserman03}, Rao-Blackwellization~\citep{Ranganath14}, importance sampling~\citep{Ruiz16, Burda16, Sakaya17,Li18}, and many others~\citep{Titsias15,Roeder17}.
Nevertheless, it is still challenging to properly construct auxiliary functions such as a control variate function and importance function in these methods.
 
After the reparameterization trick was proposed by \citet{Kingma14} and \citet{Rezende14}, many studies on the reparameterized gradient have also been conducted, e.g., studies on generalized reparameterized gradients~\citep{Ruiz2016b,tokui16,tucker17,Will18}, control variates on reparameterized gradients~\citep{Miller17,Geffner20}, and the doubly reparameterized gradient~\citep{Tucker19}.
These studies have made the reparameterized gradient estimator a more practical tool fot reducing the gradient variance in the MCVI framework.
 
 \begin{table}[t]
 \centering
 \scalebox{0.75}{
 \begin{tabular}{l|cccccc}
 \textbf{}& Method & OV& GE & SS & CA      & SNR \\ \hline \hline
 \citet{Ranganath14} & CV \& RB    & $\mathcal{O}(N^{-1})$ & SF & Stay      & -  & -   \\ 
 \citet{Ruiz16}      & IS    & $\mathcal{O}(N^{-1})$ & SF & Stay      & -  & -   \\ 
 \citet{Roeder17}    & Stop Gradient      & $\mathcal{O}(N^{-1})$ & RG & Stay      & -  & -   \\ 
 \citet{Miller17}    & CV   & $\mathcal{O}(N^{-1})$ & RG & Stay      & -  & -   \\
 \citet{Sakaya17}    & IS   & $\mathcal{O}(N^{-1})$ & SF \& RG  & Stay      & -  & -   \\ 
 \citet{Li18} & Adaptive IS & $\mathcal{O}(N^{-1})$ & SF & Stay      & -  & -   \\ 
 \citet{Buchholz18a} & RQMC & $\mathcal{O}(N^{-2})$ & SF \& RG  & Stay      & \checkmark (fixed learning rate)     & -   \\ 
 \textbf{This work}& MLMC & $\mathcal{O}(\eta_{t-1}^{2} N_{t}^{-1})$ & RG & Adaptive  & \checkmark (learning rate scheduling) & \checkmark      
\end{tabular}
}
\caption{Relationship between previous work and this work (OV: Order of variance, GE: Gradient estimator, SS: Sample size, CA: Convergence analysis, SNR: SNR analysis). “CV” stands for control variates, “RB” for Rao-Blackwellization, and “IS” for importance sampling. $N$ denotes the number of random variables from $p(\epsilon)$, and $t$ is an iteration step. “SF” means a score function, and “RG” stands for the reparameterized gradient. $N_{t}$ and $\eta_{t}$ denote the adaptively-estimated sample size and the learning rate scheduler function in iteration step $t$, respectively.
“SNR” stands for \textit{signal-to-noise} ratio.\label{Related work}}
\end{table}
{The idea of using a more sophisticated method of Monte Carlo sampling to reduce the variance of the estimator has been adopted in a Markov Chain Monte Carlo (MCMC) context, e.g., \citet{lyne15} and \citet{georgoulas17}, but has only recently been explored in the MCVI framework.}
The objective of this framework is to improve the $\mathcal{O}(N^{-1})$ rate of the gradient estimator's variance.
\citet{Ranganath14} and \citet{Ruiz16} suggested using quasi-Monte Carlo (QMC), and \citet{Tran17} applied it to a specific model.
Recently, \citet{Buchholz18a} have proposed a variance reduction method that uses randomized QMC (RQMC), called QMCVI, which can achieve, in the best case, the $\mathcal{O}(N^{-2})$ rate of the variance in the MCVI framework.
However, it is known that estimation with QMC-based methods is sometimes worse than that with MC methods because of a potentially unfavorable interaction between the underlying deterministic points and the function to be estimated~\citep{Lemi09}.

{In this paper, we focus} only on sampling-based methods such as the MC-based method~\citep{Ranganath14}, the RQMC-based method~\citep{Buchholz18a}, and our method, {and compare their theoretical properties and performance.}
The reason for this is that these methods are the core methods of gradient estimation and can be combined with the various variance reduction strategies introduced above.

\paragraph{SNR :}
SNR is a measure of the quality of a stochastic gradient estimator.
Recently, \citet{rainforth18} have analyzed the behavior of an importance-weighted stochastic gradient in terms of the SNR and revealed differences in the effect of increasing the number of importance weights between inference and generative networks on the gradient estimator in the variational autoencoder~\citep{Kingma14}.
In addition, there have been several theoretical and empirical analyses of stochastic gradient estimators using the SNR~\citep{hennig13, Lee18, shwartz17, Tucker19}.
\section{Multilevel Monte Carlo Variational Inference (MLMCVI)}
\label{mlmcvi-sec}
In this section, we introduce our proposed method, MLMCVI.
First, we explain the core idea behind proposing MLMCVI in Section~\ref{sec:idea_mlmcvi}.
Finally, we derive its inference algorithm in Section~\ref{sec:derivation}.
A summary of the MLMC method is given in Appendix~\ref{app.A}.

\subsection{Key Idea of MLMCVI}
\label{sec:idea_mlmcvi}
The key idea of MLMCVI is to construct a low-variance gradient estimator by recycling past parameters and gradients that are \emph{information obtained as the optimization proceeds.}
The equation in \ref{eq:RG_free} implies that the MLMC method is applicable to the reparameterized gradient since the expectation always depends on a fixed distribution $p(\epsilon)$, and linearity of expectation is available.
Another important idea of MLMCVI is to consider the number of levels as the number of iterations $t$ in the optimization process.
By applying these ideas, we can ``recycle'' previous parameters and {old gradient estimates.}
When we set
\begin{align}
\label{eq:g_lambda}
    {g_{\lambda_{t}}(\epsilon) =  \nabla_{\lambda_{t}} \log q (\mathcal{T}(\epsilon;\lambda_{t})|\lambda_{t}) - \nabla_{\lambda_{t}} \log p({\bf x},\mathcal{T}(\epsilon;\lambda_{t})),}
\end{align}
the Multilevel reparameterized gradient (MLRG) {at} iteration $T$ is expressed as
\begin{align}
\label{eq:MLRG}
    \nabla_{\lambda_{T}}^{\mathrm{MLRG}}\mathcal{F}(\lambda_{T}) 
    = \mathbb{E}_{p(\epsilon)}[g_{\lambda_{0}}(\epsilon)] + \sum_{t=1}^{T} \bigg(\mathbb{E}_{p(\epsilon)}[g_{\lambda_{t}}(\epsilon) - g_{\lambda_{t-1}}(\epsilon)] \bigg),
\end{align}
where $t \in \mathbb{N}$ and $T\in \mathbb{N}$.
In the MCVI framework, we need an unbiased estimator of the gradient for stochastic optimization.
An unbiased estimator for the MLRG in Eq.~\ref{eq:MLRG} can be immediately obtained as
\begin{align}
    \widehat{\nabla}_{\lambda_{T}}^{\mathrm{MLRG}}\mathcal{F}(\lambda_{T}) = N_{0}^{-1} \sum_{n=1}^{N_{0}}g_{\lambda_{0}}(\epsilon_{(n,0)}) 
    + \sum_{t=1}^{T} \bigg( N_{t}^{-1} \sum_{n=1}^{N_{t}} [g_{\lambda_{t}}(\epsilon_{(n,t)}) - g_{\lambda_{t-1}}(\epsilon_{(n,t)})] \bigg),
\end{align}
where $N_{t} \ (t=0,1,\dots,T)$ is the sample size in each iteration, and the MLRG estimator is unbiased for $\nabla_{\lambda_{T}}\mathcal{F}(\lambda_{T})$.

We note that the old gradient $g_{\lambda_{t-1}}(\epsilon_{(n,t)})$ is estimated each time using the random variables $\epsilon_{(n,t)}$ per iteration.
That is, the past gradient estimated at step $t$, i.e., $g_{\lambda_{t-1}}(\epsilon_{(n,t)})$, and the gradient estimated at step $t-1$ (which is the target at that time), i.e., $g_{\lambda_{t-1}}(\epsilon_{(n,t-1)})$, are independent and thus uncorrelated.

\subsection{Derivation}
\label{sec:derivation}
The MLRG estimator has a problem: the total cost can become crucially large as $t$ goes to infinity because it has a telescoping sum term.
To bypass this problem, we consider another formulation of the MLRG estimator and derive a special update rule on the basis of SGD {in Eq.~\ref{eq:SGD}}.

\begin{lem}[Another formulation of MLRG estimator]
\label{lem:MLRG_update}
    The MLRG estimator in iteration $t \geq 1$ can be represented as
    \begin{align*}
        \widehat{\nabla}_{\lambda_{t}}^{\mathrm{MLRG}} \mathcal{F}(\lambda_{t}) = \widehat{\nabla}_{\lambda_{t-1}}^{\mathrm{MLRG}}\mathcal{F}(\lambda_{t-1}) 
                + N_{t}^{-1} \sum_{n=1}^{N_{t}} \bigg[ g_{\lambda_{t}}(\epsilon_{(n,t)}) - g_{\lambda_{t-1}}(\epsilon_{(n,t)}) \bigg],
    \end{align*}
    and the update rule for this estimator in SGD is
    \begin{align*}
        \lambda_{t+1} = \lambda_{t} + \frac{\eta_{t}}{\eta_{t-1}} (\lambda_{t} - \lambda_{t-1})
        - \alpha_{t} N_{t}^{-1} \sum_{n=1}^{N_{t}} \bigg[g_{\lambda_{t}}(\epsilon_{(n,t)}) - g_{\lambda_{t-1}}(\epsilon_{(n,t)}) \bigg].
    \end{align*}
    \begin{proof}
    See Appendix \ref{lem1:proof} for the proof.
    \end{proof}
\end{lem}
From this result, we can obtain the MLRG estimator by simply saving the previous parameters and computing the two gradients.

How can we estimate the optimal sample size $N_{t}$ of the MLRG estimator at each $t \ (t \geq 1)$?
In the context of variance reduction, we can consider the optimal sample size $N_{t}$ as the one that minimizes the \emph{total} variance of the MLRG estimator.
{In fact, as in the study by \citet{Giles15}, the optimal $N_{t}$ can be derived by temporarily treating $N_{t}$ as a real number and solving the constrained optimization problem.
To derive this, we set the following definition and assumption.}
\begin{definition}[One-sample cost and variance]
\label{def:comp_var}
Let $C_{t}$ and $\mathbb{V}_{t}$ be the one-sample  computational cost and variance of $g_{\lambda_{t}}(\epsilon_{(n,t)}) - g_{\lambda_{t-1}}(\epsilon_{(n,t)})$ {at} iteration $t$, respectively.
{At iteration $t=0$, let $C_{0}$ and $\mathbb{V}_{0}$ be the one-sample computational cost and variance of $g_{\lambda_{0}}(\epsilon_{(n,0)})$, respectively.}
\end{definition}
{From the above definition, the total cost and variance of $\widehat{\nabla}_{\lambda_{T}}^{\mathrm{MLRG}}\mathcal{F}(\lambda_{T})$ $(T \geq 1)$ can be expressed as {$\sum_{t=0}^{T}N_{t}C_{t}$ and $\sum_{t=0}^{T}N_{t}^{-1}\mathbb{V}_{t}$, respectively.}
Since the computational cost of the gradient calculation in {each iteration of SGD is} $\mathcal{O}(d)$~\citep{Mu17}, where $d$ is {the dimension of the parameter space}, i.e., $\lambda \in \mathbb{R}^{d}$, we can see that $C_{0} \leq \nu d$ and $C_{t} \leq 2 \nu d$ ($t \geq 1$), where $\nu$ is a positive constant.
That is, the computational cost of our method does not depend on the iteration $t$.
Throughout this paper, we assume that this upper bound is the actual computational cost as follows.}
\begin{assump}[Computational cost {of} gradient calculation]
\label{assmp:comp_cost}
The one-sample computational costs of $g_{\lambda_{0}}(\epsilon_{(n,0)})$ and $g_{\lambda_{t}}(\epsilon_{(n,t)}) - g_{\lambda_{t-1}}(\epsilon_{(n,t)})$ are
$C_{0} = \nu d$ and $C_{t} = 2 \nu d$ $(t \geq 1)$, respectively.
\end{assump}
Assumption~\ref{assmp:comp_cost} {corresponds to a} ``pessimistic'' case for the computational cost of gradient estimation in each iteration.
Under Definition~\ref{def:comp_var} and Assumption~\ref{assmp:comp_cost}, we can establish the following theorem according to standard proof techniques reported by \citet{Giles08}.

\begin{thm}[Optimal sample size $N_{t}$]
\label{thm:optimal_sample}
Suppose that Assumption~\ref{assmp:comp_cost} holds.
{Let $N_{t}$ be a positive real number for all $t$}. 
Suppose that a total computational cost for sampling is fixed, i.e., {$\sum_{t=0}^{T} N_{t}C_{t} \leq \sum_{t=0}^{T} C_{t} N_{0}$}.
Then, the optimal sample size for $t \geq 1$, which minimizes the overall variance {$\sum_{t=0}^{T}N_{t}^{-1}\mathbb{V}_{t}$}, is
\begin{align}
    \label{estimate_num_ceil}
    &N_{t+1} = \sqrt{\frac{\mathbb{V}_{t+1}}{\mathbb{V}_{t}}} N_{t},
\end{align}
\diff{
where $N_{1} = \frac{(2T+1)\sqrt{\mathbb{V}_{1}}}{\sqrt{2\mathbb{V}_{0}} + 2 \sum_{t=1}^{T} \sqrt{\mathbb{V}_{t}}}N_{0}$.}
\begin{proof}(Sketch of Proof)
This theorem can be proved by solving a constrained optimization problem
that minimizes the overall variance as {$\min_{N_{t}} \sum_{t=0}^{T} N_{t}^{-1} \mathbb{V}_{t} \ \text{s.t.} \ \sum_{t=0}^{T} N_{t}C_{t} \leq \sum_{t=0}^{T} C_{t} N_{0}$.}
By solving this for $t \geq 1$, we can obtain $N_{t} = \frac{1}{\mu} \sqrt{\frac{\mathbb{V}_{t}}{C_{t}}}$, where {$\mu > 0$}. 
\diff{Taking the ratio $N_{t+1}/N_{t}$ leads to the above result.}
The complete proof is given in Section~\ref{thm1:proof}.
\end{proof}
\end{thm}
\diff{The constrained optimization problem above implies that we construct the optimal sample size $N_{t}$ so as to minimize the total variance at a lower cost than when the sample size is fixed as the initial sample size $N_{0}$.}
Furthermore, Theorem~\ref{thm:optimal_sample} shows that the optimal sample size $N_{t+1}$ does not depend on the cost of computing the gradients when $N_{t}$ is given but depends on the ratio of the \emph{one-sample} variances.

Unfortunately, we still cannot estimate the optimal $N_{t}$ using Eq.~\ref{estimate_num_ceil} because the true $\mathbb{V}_{t+1}$ and $\mathbb{V}_{t}$ are unknown.
To bypass this problem, we consider an alternative way by minimizing {an upper bound on the variance.}
Therefore, the magnitude of $\mathbb{V}_{t}$ is critical for sample size estimation.
To confirm this perspective, we analyze the \emph{one-sample} gradient variance under the following assumptions, which are often considered in the MCVI context.
\begin{assump}[Variational distribution in a multivariate location-scale family]
\label{asm:gaussian}
The variational distribution $q(\mathcal{T}(\epsilon;\lambda)|\lambda)$ is in a multivariate location-scale family with a single vector of {parameters} $\lambda = ({\bf m}, {\bf v})$.
\end{assump}
\begin{assump}[Boundedness of $\mathcal{T}(\epsilon;\lambda)$]
\label{asm:lipshitz2}
The reparameterized random variable $\mathcal{T}(\epsilon;\lambda)$ is bounded, i.e., $\exists K_{1} > 0 \ \textrm{s.t.} \ \forall \lambda$: $\|\mathcal{T}(\epsilon;\lambda)\|_{2}^{2} \leq K_{1}$.
\end{assump}
\begin{assump}[Robbins--Monro condition]
\label{asm:robbins_monro}
The learning rate $\alpha_{t}$ satisfies the following conditions:
$\sum_{t=0}^{\infty}\alpha_{t} = \infty$ and $\sum_{t=0}^{\infty}\alpha_{t}^{2} < \infty$.
\end{assump}
Assumption \ref{asm:gaussian} has been extensively used in several VI frameworks with a stochastic gradient~\citep{Kingma14,Rezende14}.
In Assumption \ref{asm:lipshitz2}, we also assume that $\mathcal{T}(\epsilon;\lambda)$ is bounded.
Although this assumption is not always {satisfied}, we can make it hold by alternative techniques, e.g., by truncating $\mathcal{T}(\epsilon;\lambda)$ to a particular value by the proximal operator~\citep{nesterov83}.
The details and the new algorithm obtained from this truncation are described in Appendix~\ref{app.non-bounded}.
The Robbins--Monro condition (Assumption~\ref{asm:robbins_monro}) is often assumed in convergence analysis on SGD with learning rate scheduling~\citep{bottou16}.

Under {the assumptions above}, we give the following proposition for the \emph{one-sample} gradient variance.
\begin{prop}[Order of one-sample gradient variance]
\label{prop:one-sample_variance}
Suppose that Assumptions \ref{asm:gaussian}--\ref{asm:robbins_monro} hold.
Then, the expectation of the $l_{2}$-norm of $g_{\lambda_{t}}(\epsilon) - g_{\lambda_{t-1}}(\epsilon)$ in iteration $t \ (t \geq 1)$ is bounded as
\begin{align*}
    \mathbb{E}_{p(\epsilon)} \bigg[ \| g_{\lambda_{t}}(\epsilon) - g_{\lambda_{t-1}}(\epsilon) \|_{2}^{2} \bigg] \leq \eta_{t-1}^{2} K_{1} (c_{1} + d \delta c_{2}),
\end{align*}
where $\delta,c_{1},c_{2}$ are positive constants.
\begin{proof}
See Appendix~\ref{prop1:proof} for the proof.
\end{proof}
\end{prop}
According to Proposition~\ref{prop:one-sample_variance}, the order of the \emph{one-sample} variance $\mathbb{V}_{t}$ is $\mathcal{O}(\eta_{t-1}^{2})$; therefore, $\mathbb{V}_{t} \overset{t \rightarrow \infty}{\rightarrow} 0$.
From the above results, we can derive an alternative way for sample size estimation as follows.
\begin{thm}[Alternative sample size estimation]
\label{thm:alt_optimal_sample}
Suppose that Assumptions~\ref{assmp:comp_cost}--\ref{asm:robbins_monro} hold and a total computational cost for sampling is fixed, i.e., {$\sum_{t=0}^{T} N_{t}C_{t} \leq \sum_{t=0}^{T} C_{t} N_{0}$}.
{Let $N_{t}$ be a positive real number for all $t$}. 
Then, the optimal sample size for $t \geq 1$, which minimizes the upper bound of the total variance, is
\begin{align}
\label{estimate_num_ceil_another}
    N_{t+1} = \eta_{t}N_{1},
\end{align}
where $\kappa$ is a positive constant and \diff{$\frac{2T + 1}{2T + \sqrt{2}}N_{0} \leq N_{1} \leq N_{0}$.}
\begin{proof}
See Section~\ref{thm2:proof} for the proof.
\end{proof}
\end{thm}
\diff{
\begin{rem}
From Theorems~\ref{thm:optimal_sample} and \ref{thm:alt_optimal_sample}, since the optimal $N_{t}$ is derived at $t \geq 1$ and $N_{0}$ is given as an initial value, we cannot estimate $N_{1}$ by using the ``optimal'' $N_{0}$.
Therefore, we have to set $N_{1}$ so that the following condition holds: $\frac{2T + 1}{2T + \sqrt{2}}N_{0} \leq N_{1} \leq N_{0}$.
From this, we can see $N_{1} \simeq N_{0}$ if we set $T$ to be large enough.
When $T=100$ and $N_{0}=100$, for example, we can see $99.2978 \simeq \frac{201}{200 + \sqrt{2}} \cdot 100 \leq N_{1} \leq 100$. 
Therefore, even with the assumption that $N_{0} = N_{1}$, the impact on our algorithm can be viewed as small.
From now on, we discuss our method assuming that $N_{0} = N_{1}$.
\end{rem}}
From Theorem~\ref{thm:alt_optimal_sample}, we can estimate the sample size {by using only the learning rate} scheduler $\eta_{t}$.
Since the sample size $N_{t}$ has to be a positive integer, we use the ceiling function in Eq.~\ref{estimate_num_ceil_another}, i.e., $N_{t+1}^{*} = \lceil{N_{t+1} \rceil} = \lceil{\eta_{t}N_{1}\rceil}$, where $\lceil{x \rceil} = \min \{k \in \mathbb{Z} | x \leq k \}$.
\diff{Although the cost constraint condition of Theorem~\ref{thm:alt_optimal_sample} seems to be violated by rounding $N_{t}$ via the ceiling function, we show that this need not be a concern if we set $\eta_{t}$ so that the following definition holds.
\begin{definition}[Low total cost condition]
\label{def:low_total_cost_condition}
 We call the low total cost condition if the total cost added by $r_{t}$ for each $t$ satisfies the following condition:
 \begin{align*}
  \sum_{t=0}^{T}  (N_{t} + r_{t})C_{t} < \sum_{t=0}^{T} C_{t}N_{0},
\end{align*}
where $r_{t}$ is a positive real number satisfying $0 < r_{t} < 1$ and $r_{0} = 0$ for all $t$.
\end{definition}
The above condition implies that the total cost added by $r_{t}$ for all $t \ (t \geq 1)$ does not exceed the total cost constraint in Theorem~\ref{thm:alt_optimal_sample}.
Now we show the sufficient condition of the learning scheduler function $\eta_{t}$ to satisfy the low total cost condition in Definition~\ref{def:low_total_cost_condition} as follows.
}
\begin{algorithm}[t]                      
                \caption{Multilevel Monte Carlo Variational Inference}
                \label{alg3}
                \begin{algorithmic}[1]
                        \REQUIRE {Data ${\bf x}$, random variable $\epsilon \sim p(\epsilon)$, transform ${\bf z} = \mathcal{T}(\epsilon;\lambda)$, model $p({\bf x},{\bf z})$, 
                        variational family $q({\bf z}|\lambda)$}
                        \ENSURE {Variational parameter $\lambda^{*}$}
                        \STATE \textbf{Initialize:} $N_{0}$, $\lambda_{0}$, $\alpha_{0}$, and hyperparameter of $\eta$
                        \FOR {$t = 0$ to $T$}
                        \IF {$t=0$}
                        \STATE $\epsilon_{n} \sim p(\epsilon) \ (n = 1,2,\dots,N_{0})$ \ $\triangleleft$ \ \textrm{sampling} $\epsilon$
                        \STATE $\hat{g}_{\lambda_{0}}(\epsilon_{1:N_{0}^{*}}) = N_{0}^{-1} \sum_{n=1}^{N_{0}} g_{\lambda_{0}}(\epsilon_{n})$ \ $\triangleleft$ \ \textrm{calc. RG estimator}
                        \STATE $\lambda_{1} = \lambda_{0} - \alpha_{0} \hat{g}_{\lambda_{0}}(\epsilon_{1:N_{0}})$ \ $\triangleleft$ \ \textrm{grad-update}
                        \ELSE
                        \STATE \textbf{estimate} $N_{t}$ \textrm{using} $N_{t} = \lceil{\eta_{t-1} N_{0} \rceil}$
                        \STATE $\epsilon_{n} \sim p(\epsilon) \ (n = 1,2,\dots,N_{t})$ \ $\triangleleft$ \ \textrm{sampling} $\epsilon$
                        \STATE $\hat{g}^{'}_{\lambda_{t}}(\epsilon_{1:N_{t}}) = N_{t}^{-1} \sum_{n=1}^{N_{t}} [g_{\lambda_{t}}(\epsilon_{(n,t)}) - g_{\lambda_{t-1}}(\epsilon_{(n,t)}) ]$
                        \ $\triangleleft$ \ \textrm{calc. Multilevel term}
                        \STATE $\lambda_{t+1} = \lambda_{t} + \frac{\eta_{t}}{\eta_{t-1}}(\lambda_{t} - \lambda_{t-1}) - \alpha_{t} \hat{g}^{'}_{\lambda_{t}}(\epsilon_{1:N_{t}})$ \ $\triangleleft$ \ \textrm{grad-update}
                        \IF {$\lambda_{t+1}$ \textrm{has converged to} $\lambda^{*}$}
                        \STATE \textbf{break}
                        \ENDIF
                        \ENDIF
                        \ENDFOR
                        \RETURN $\lambda^{*}$
                \end{algorithmic}
        \end{algorithm}
\diff{
\begin{lem}[Sufficient conditions for satisfying low total cost condition]
\label{cor:optimal}
Suppose \\ that Assumption~\ref{assmp:comp_cost} holds.
Let $N_{t}$ be a positive real number for all $t$ and $N_{t+1}$ be estimated as $N_{t+1} = \eta_{t}N_{1}$ for $t \geq 1$ where $N_{1} = N_{0}$.
\diff{
If we set $\eta_{t}$ to be $\eta_{t} \leq 1 - \frac{2}{N_{0}}$ at least after the iteration $\lfloor \frac{T}{2} \rfloor$ $(T \geq 2)$, where $\lfloor x \rfloor = \max \{k \in \mathbb{Z}| k \leq x \}$, the low total cost condition holds.
}
\begin{proof}
See Appendix~\ref{cor1:proof} for the proof.
\end{proof}
\end{lem}}
\diff{
This lemma implies that, if we set the learning rate scheduler function to be less than $1 - \frac{2}{N_{0}}$ after at least about half of the iteration $T$, the original constraint condition in Theorem~\ref{thm:alt_optimal_sample} is not violated even if the total cost is increased by $0 < r_{t} < 1$ for all $t$.
Namely, the optimal $N_{t}$ is invariant under the above condition even if the original total cost is increased by the ceiling function.
Let us consider using the step-based decay function (see Appendix~\ref{sec:another_samplesize}) and setting $N_{0}=100$, $T=1000$, and $\{\beta, r\} = \{0.5, 100 \}$ as an example.
Then, the function $\eta_{t}$ needs to be less than $1 - \frac{2}{N_{0}} = 0.998$ at least in iteration $t = \lfloor \frac{1000}{2} \rfloor= 500$.
We can see $\eta_{500} = \beta^{\lfloor \frac{500}{r}  \rfloor} = 0.5^{\lfloor \frac{500}{100}  \rfloor} = 0.03125 < 0.998$ and the condition holds.
To guarantee the result of Lemma~\ref{cor:optimal}, we have to assume that $T \geq 2$.
However, in MCVI, we often set a sufficiently large $T$ until the optimization converges; therefore, this assumption is quite natural.}
With this property, if we set the hyperparameter of $\eta_{t}$ properly, we can use the ceiling function for the optimal sample size derived in Theorem~\ref{thm:alt_optimal_sample}, regardless of the increase in the cost constraint condition. 
For simplicity, we denote $N_{t}^{*}$ as $N_{t}$ hereafter.
\diff{Furthermore, since we assume that $N_{0} = N_{1}$, we can rewrite the sample size estimation scheme as $N_{t} = \lceil \eta_{t-1}N_{0} \rceil$, where we denote $\eta_{-1} = \eta_{0} = 1$.}
Straightforwardly, we can see that $N_{t}$ goes to $1$ as $t \rightarrow \infty$ because $\eta_{t-1} \rightarrow 0$.
Therefore, our method can reduce the sample cost for gradient estimation as the optimization proceeds.

According to Lemmas~\ref{lem:MLRG_update}--\ref{cor:optimal} and Theorem~\ref{thm:alt_optimal_sample}, the MLMCVI algorithm is derived as Algorithm~\ref{alg3}.

\section{Theoretical Analyses}
\label{sec:theoretical_analysis}
In this section, we analyze the effect of our method on the basis of the weighted average norm of the gradient and SNR.
In addition, we compare the results with sampling-based methods such as MCVI and QMCVI~\citep{Buchholz18a}.
For this analysis, we add the following assumption, which is often considered in the MCVI context.
\begin{assump}[Lipschitz continuity on $\nabla_{\lambda}\mathcal{F}(\lambda)$]
\label{asm:lipschitz}
{The variational free energy} $\mathcal{F}(\lambda)$ is a function with Lipschitz continuous derivatives, i.e., $\exists K_{2} > 0$ s.t. $\forall \lambda, \bar{\lambda}$: $\|\nabla_{\lambda}\mathcal{F}(\lambda) - \nabla_{\bar{\lambda}}\mathcal{F}(\bar{\lambda})  \|_{2}^{2} \leq K_{2} \|\lambda - \bar{\lambda} \|_{2}^{2}$.
\end{assump}
This assumption means that the gradient of {the variational free energy} cannot change very rapidly as $\lambda$ changes~\citep{Buchholz18a,Domke19}.


\subsection{Convergence Analysis}
\label{subsec:conv_anal}
{Here, we first analyze the convergence of existing methods and our method theoretically and compare them.}
In our method, since the MLRG estimator appears after the first optimization step, we focus on the case where the number of iterations $t$ is $t \geq 1$.
We should first analyze the order of $\mathbb{V}[ \widehat{\nabla}_{\lambda_{t}}^{\textrm{MLRG}} \mathcal{F}(\lambda_{t}) ]$ so that we can perform the convergence analysis because it has been shown by \citet{Buchholz18a} that the convergence of MCVI depends on the variance of the gradient estimator.
\begin{lem}[Variance of $\widehat{\nabla}_{\lambda_{t}}^{\textrm{MLRG}} \mathcal{F}(\lambda_{t})$]
\label{lem:var_order}
Suppose that Assumptions~\ref{assmp:comp_cost}--\ref{asm:robbins_monro} hold.
Then, the order of $\mathbb{V}[ \widehat{\nabla}_{\lambda_{t}}^{\textrm{MLRG}} \mathcal{F}(\lambda_{t}) ]$ is $\mathcal{O}(\eta_{t-1}^{2} N_{t}^{-1})$.
\begin{proof}
See Section \ref{lem3:proof} for the proof.
\end{proof}
\end{lem}
From the results from Theorem~\ref{thm:alt_optimal_sample} and Lemma~\ref{lem:var_order}, and the fact that $\eta_{t} \rightarrow 0$ as $t \rightarrow \infty$, we can see that the variance of our method, $\mathbb{V}[ \widehat{\nabla}_{\lambda_{t}}^{\textrm{MLRG}} \mathcal{F}(\lambda_{t}) ]$, converges to 0.

Regarding stochastic optimization, \citet{bottou16} provided comprehensive theorems on the basis of SGD.
On the basis of those theorems, we can prove the following upper bounds on the norm of the gradient and use them to analyze the effect of our method.
We can also compare the upper bounds with those for the MC-, RQMC-, and MLMC-based methods.
Before showing the results, for simplification, we define the common terms in the other expressions as
\begin{align*}
    G_{T} = \frac{1}{A_{T}} [\mathcal{F}(\lambda_{1}) - \mathcal{F}(\lambda^{*})] + \frac{\alpha_{0}^{2} K_{1}}{2A_{T}} \sum_{t=1}^{T} \eta_{t}^{2} \mathbb{E}\bigg[||\nabla_{\lambda_{t}} \mathcal{F}(\lambda_{t})||_{2}^{2}\bigg],
\end{align*}
where $A_{T} = \sum_{t=1}^{T}\alpha_{t}$, $\lambda^{*}$ is the optimal parameter and $\lambda_{t}$ is iteratively defined in the SGD update rule.
\begin{thm}[Weighted average norm of gradient]
\label{thm:norm_MC,RQMC}
Suppose that Assumptions~\ref{assmp:comp_cost}--\ref{asm:lipschitz} hold.
Then, in each of the MC-, RQMC-, and MLMC-based methods, the weighted average norm of the gradient at iteration $t \geq 1$ is bounded as
\begin{align*}
            \frac{1}{A_{T}}\sum_{t=1}^{T} \alpha_{t}\mathbb{E}\bigg[||\nabla_{\lambda_{t}} \mathcal{F}(\lambda_{t})||_{2}^{2} \bigg] \leq 
            \begin{cases}
            G_{T} + \frac{\alpha_{0}^{2}K_{1}}{2A_{T}}\sum_{t=1}^{T} \eta_{t}^{2} {\frac{\kappa}{N}} \  \mathrm{(MC)}, \\
            G_{T} + \frac{\alpha_{0}^{2}K_{1}}{2A_{T}}\sum_{t=1}^{T}  \eta_{t}^{2} \frac{\kappa}{N^{2}} \mathrm{(RQMC)}, \\
            G_{T} + \frac{\alpha_{0}^{2} K_{1}}{2A_{T}} \sum_{t=1}^{T} \eta_{t}^{2}\eta_{t-1}^{2} \frac{\kappa}{N_{t}} \mathrm{(MLMC)},
            \end{cases}
\end{align*}
respectively.
\begin{proof}
See Appendix \ref{thm3:proof} and \ref{thm4:proof} for the proofs.
\end{proof}
\end{thm}
Theorem \ref{thm:norm_MC,RQMC} states that the weighted average norm of the squared gradients converges to zero because of $A_{T} = \sum_{t=1}^{T}\alpha_{t} = \infty$ and Assumption \ref{asm:robbins_monro} even if the gradient estimator is noisy.
This fact can guarantee that the expectation of the gradient norms of the MC-, RQMC-, and MLMC-based methods asymptotically remains around zero.
In addition, the difference in convergence speed between these methods depends on the last term in each of these bounds.
While the convergence of the MC- and RQMC-based methods can only be accelerated by increasing the number of samples, i.e., by increasing the gradient estimation cost, our method can be accelerated without increasing the estimation cost (rather, while reducing it since $N_{t} \rightarrow 1$ as $t \rightarrow \infty$; see Section~\ref{sec:derivation}) using the learning rate {scheduler} function $\eta_{t-1}^{2}$.

\subsection{SNR Analysis}
SNR is a useful tool to confirm the behavior of gradient estimation.
Recently, for example, \citet{rainforth18} have been analyzed using SNR the gradient behavior of various importance-weighted autoencoder.
In the context of variance reduction, it is also important to investigate how our method affects gradient estimation and how it changes compared with existing methods to theoretically guarantee its performance.
Therefore, we evaluate the quality of MC-, RQMC- and MLMC-based gradient estimators through the SNR analysis.

The SNR of a gradient estimator is defined as
\begin{align*}
 \mathrm{SNR}(\lambda) = \frac{\|\mathbb{E}_{p(\epsilon_{1:N})}[\hat{g}_{\lambda}(\epsilon_{1:N})]] \|_{2}^{2}}{\sqrt{\mathbb{V}[\hat{g}_{\lambda}(\epsilon_{1:N})]}},
\end{align*}
where $\hat{g}_{\lambda}(\epsilon_{1:N})$ is given by Eq.~\ref{eq:g_lambda_estimator}.
This indicates that if $\mathrm{SNR} \rightarrow 0$, the gradient estimator is dominated by random noise, which degrades the accuracy of estimation.

On the basis of the theoretical properties from Lemma~\ref{lem:var_order}, we prove the following theorem for SNR.
\begin{thm}[Signal-to-noise ratio bound]
\label{prop:SNR}
Suppose that Assumptions~\ref{assmp:comp_cost}--\ref{asm:lipschitz} hold.
Then, the SNR {at} iteration $t$ for each method is bounded as
\begin{align*}
    \mathrm{SNR}(\lambda_{t}) \geq
    \begin{cases}
      \frac{\| \nabla_{\lambda_{t}}\mathcal{F}(\lambda_{t}) \|_{2}^{2}}{\sqrt{\kappa}} \cdot \sqrt{N} & (\mathrm{MC}), \\
    \frac{\| \nabla_{\lambda_{t}}\mathcal{F}(\lambda_{t}) \|_{2}^{2}}{\sqrt{\kappa}} \cdot N & (\mathrm{RQMC}), \\
      \frac{\|\nabla_{\lambda_{t}}\mathcal{F}(\lambda_{t})  \|_{2}^{2}}{\sqrt{\kappa}} \cdot \frac{\sqrt{N_{t}}}{\eta_{t-1}} & (\mathrm{MLMC}),
    \end{cases}
\end{align*}
where $\kappa$ is a positive constant and $N$ is an initial sample size for the MC- and RQMC-based methods.
\begin{proof}
See Section \ref{prop2:proof} for the proof.
\end{proof}
\end{thm}
The above theorem implies that for the MC- and RQMC-based methods, the SNR can be increased only by increasing the initial sample size, i.e., by increasing the estimation cost.
Furthermore, the SNRs of these methods gradually decrease because the sample size $N$ is fixed and $||\nabla_{\lambda_{t}}\mathcal{F}(\lambda_{t}) ||_{2}^{2}$ approaches $0$ as the optimization proceeds.
This fact means that the gradient estimator based on these methods becomes dominated by random noise.
On the other hand, in our method, it seems that the SNR gradually becomes lower than those of the MC- and RQMC-based methods since the sample size $N_{t}$ decreases as shown in Section~\ref{sec:derivation}.
However, it can be seen that our method can keep it at a high value since the learning rate scheduler function $\eta_{t-1}$ {appears in the denominator}.
That is, our method can improve the quality of gradient estimation by the learning rate scheduler function, i.e., not by increasing the estimation cost.
\section{Experiments}
\label{exp-sec}
In this section, we carried out experiments to analyze the optimization and prediction performance of our method using three models, which have become benchmark experiments in the context of variance reduction: hierarchical linear regression (HLR), Bayesian logistic regression (BLR), and Bayesian neural network (BNN) regression~\citep{Miller17,Buchholz18a}, and we compared the results with those of existing methods.
First, we compared {the optimization performance} using ELBO and log-likelihood for the training and test datasets.
Note that the reason for using ELBO in the experimental results is for consistency with the experimental setup of related studies (e.g., \cite{Miller17} and \cite{Buchholz18a}), and that ELBO and variational free energy are not essentially different. Second, we compared the performance of variance reduction by measuring the empirical gradient variance.
Finally, we determined the quality of the gradient estimator on the basis of the empirical SNR.
Throughout all experiments, we used Algorithm~\ref{alg3} for our method.

\subsection{Experiments on Benchmark Dataset}
\label{subsec:exp_benchmark}
Here, we report the results of experiments conducted to validate the effectiveness of our method.

\subsubsection{Model setting and benchmark dataset}
In our experiments, we used three different settings: HLR, BLR, and BNN.
A brief summary of the model settings and benchmark datasets used in the experiment is as follows.
\paragraph{Hierarchical Linear Regression (HLR):}
We applied HLR to toy data generated from the same generation process of model setting.
Then, the dimension of the entire parameter space was $d = 1012$. 
\paragraph{Bayesian Logistic Regression (BLR):}
We applied BLR to the breast cancer dataset in the UCI Machine Learning Repository\footnote[1]{{\url{https://archive.ics.uci.edu/ml/datasets/Breast+Cancer}}}.
Then, the dimension of the entire parameter space was $d = 62$.
\paragraph{Bayesian Neural Network (BNN):}
We applied BNN regression to the wine-quality-red dataset, which is included in the wine-quality dataset in the UCI Machine Learning Repository\footnote[2]{{\url{https://archive.ics.uci.edu/ml/datasets/Wine+Quality}}}.
Then, the dimension of the entire parameter space is $d = 653$.

We approximated these models using a variational diagonal Gaussian distribution.
A more detailed description of the settings for each model is in Appendices~\ref{hbr_setting}, \ref{blr_setting}, and \ref{bnnreg_setting}.
\begin{figure}[t]
    \centering
    \includegraphics[scale=0.28]{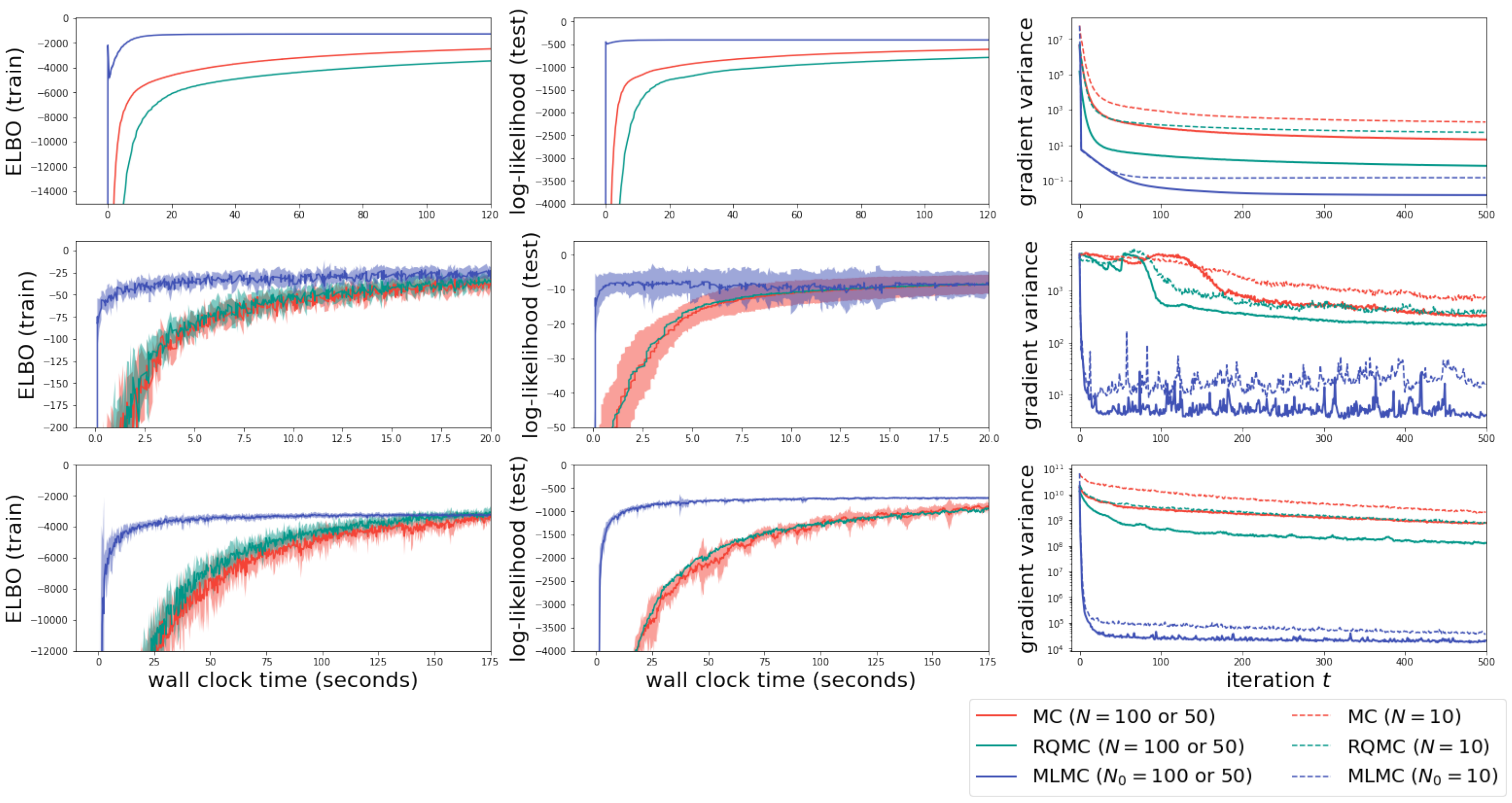}
    \caption{Experimental results. From left to right, the graphs show training ELBO (higher is better; mean $\pm$ std), test log-likelihood (higher is better; mean $\pm$ std), and empirical variance (lower is better). From top to bottom, results for HLR, BLR, and BNN.}
    \label{fig:exp_results}
\end{figure}
\begin{figure}[t]
    \centering
    \includegraphics[scale=0.29]{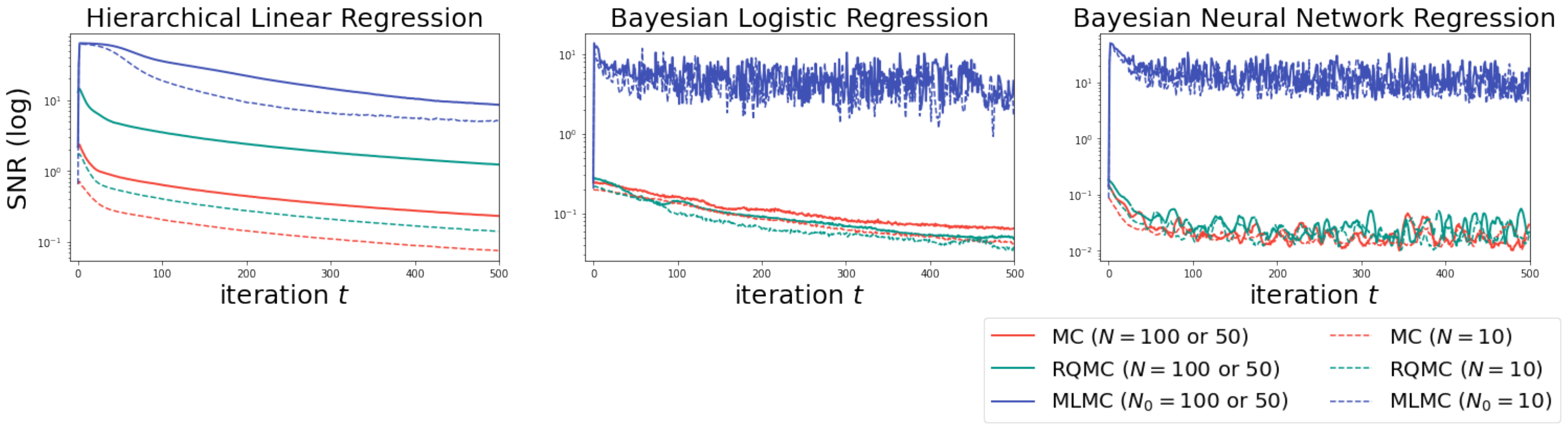}
    \caption{Experimental results on empirical SNR (higher is better).}
    \label{fig:snr}
\end{figure}
\begin{figure}[t]
    \centering
    \includegraphics[scale=0.40]{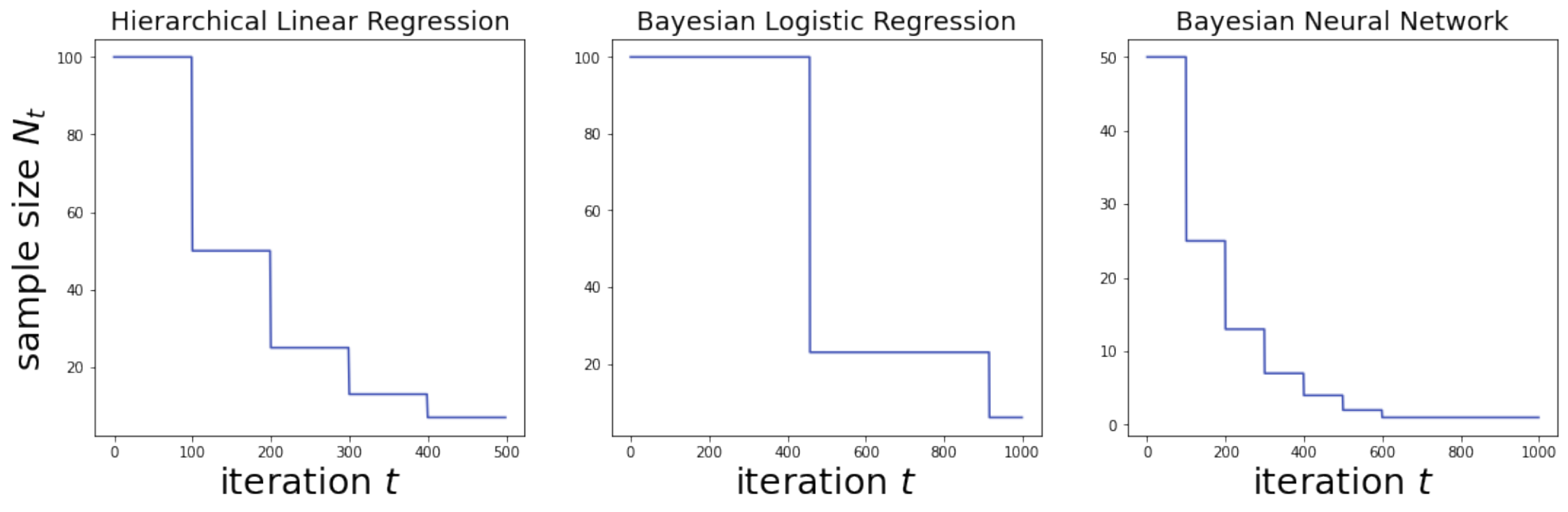}
    \caption{Behavior of estimated sample size.}
    \label{fig:sample_size}
\end{figure}

\subsubsection{Experimental Settings}
Two baseline methods were used in the experiments: vanilla MCVI based on MC sampling and QMCVI based on RQMC sampling~\citep{Buchholz18a}.
As described by \cite{Buchholz18a}, the RQMC samples were generated using the {\sf{R}} package {\sf{randtoolbox}}\footnote[3]{{\url{https://cran.r-project.org/web/packages/randtoolbox/index.html}}}.

In the experiments, the parameters $\lambda$ were initialized to be of the same values for each method.
For the optimization of variational free energy, we used Adam for the MC- and RQMC-based methods and the SGD optimizer with the learning rate scheduler $\eta$ for our method.
Furthermore, we adopted a step-based decay function as the learning rate scheduler.
The initial learning rate and the hyperparameters of the learning rate scheduler (i.e., $\beta$ and $r$) were optimized through the Tree-structured Parzen Estimator (TPE) sampler in {\sf{optuna}}\footnote[4]{{\sf optuna} is a hyperparameter optimization framework introduced by~\citet{optuna_2019} from Preferred Networks, Inc.
Thanks to \emph{define-by-run API}, we are allowed to construct the parameter search space dynamically and flexibly. {\url{https://optuna.org/}}}~\citep{optuna_2019} in $50$ trials, where $\beta$ and $r$ are the decay and drop-rate parameters, respectively.
The selected parameters are summarized in Appendix~\ref{app:additional_exp_results} (Table~\ref{tab:hyper}).
In all experiments, the training ELBO and the test log-likelihood were estimated with $2000$ MC samples.
The gradient variance was estimated by resampling the MC sample and calculating the variance of the gradient $1000$ times and computing the empirical variance at each optimization step.
In each experiment, the dataset was randomly divided into training data and test data at the ratio of $8$:$2$.
Each experiment was repeated $10$ times.

\subsubsection{Results}
\label{results}
The main experimental results are shown in Figure~\ref{fig:exp_results}.
These results show that our method (solid blue line) converged faster than the baseline methods did in all settings.
Furthermore, the results of gradient variance (rightmost column {in Figure}~\ref{fig:exp_results}) confirm that our method showed an empirically lower variance gradient estimate than the baseline methods.

The results on empirical SNR are shown in Figure~\ref{fig:snr}.
These results show that the empirical SNR of our method (dashed line and solid blue line) is higher than those of the baseline methods, corresponding to the empirical variance.
This implies that our method provides better gradient estimation than the baseline method in terms of SNR.

Furthermore, we show the behavior of the estimated sample size $N_{t}$ in our method in Figure~\ref{fig:sample_size}.
The results show that our method achieves faster convergence while reducing the sample size for gradient estimation.

As we mentioned regarding several theoretical analyses in Section~\ref{sec:theoretical_analysis}, the optimization performance of our method depends on the learning rate scheduler function $\eta$.
To understand the characteristics of our method in more detail, we conducted benchmark experiments with various hyperparameter settings and initial learning rates to determine how the performance of the optimization changes and compare it with those with the optimal hyperparameter settings.
The results are shown in the Appendix~\ref{app:additional_exp_results}.
From these results, we confirmed that, in practice, our method requires the careful optimization of the hyperparameters, especially the drop-rate $r$, and the initial learning rate.

\subsection{Bayesian Logistic Regression for Multiclass Classification}
\label{blr_setting_multi}
To confirm the performance using a more realistic problem setting and dataset, we conducted experiments on a multiclass classification task for a large image dataset such as the Fashion-MNIST dataset.

\subsubsection{Experimental settings and Results}
\label{exp:multi}
We used {Bayesian logistic regression} in this experiment.
The model settings are described in Appendix~\ref{blr_setting}.
In these settings, the dimension of the entire parameter space is $d = 7852$, and this model was also approximated by a variational diagonal Gaussian distribution.
To analyze only the effect of variance reduction and to fairly compare our method with the baseline methods, the variational free energy was optimized using the SGD optimizer with the learning rate scheduler function $\eta$ for all methods.
We used $100$ initial MC or RQMC samples for gradient estimation.
In the optimization step, we used $\eta$ as the step-decay function and set the hyperparameter $\{\beta, r\}$ for sample size estimation to $\{0.5, 100 \}$.
Finally, we set the initial learning rate as $0.01$, $0.005$, or $0.001$.

\begin{figure}[t]
    \centering
    \includegraphics[scale=0.29]{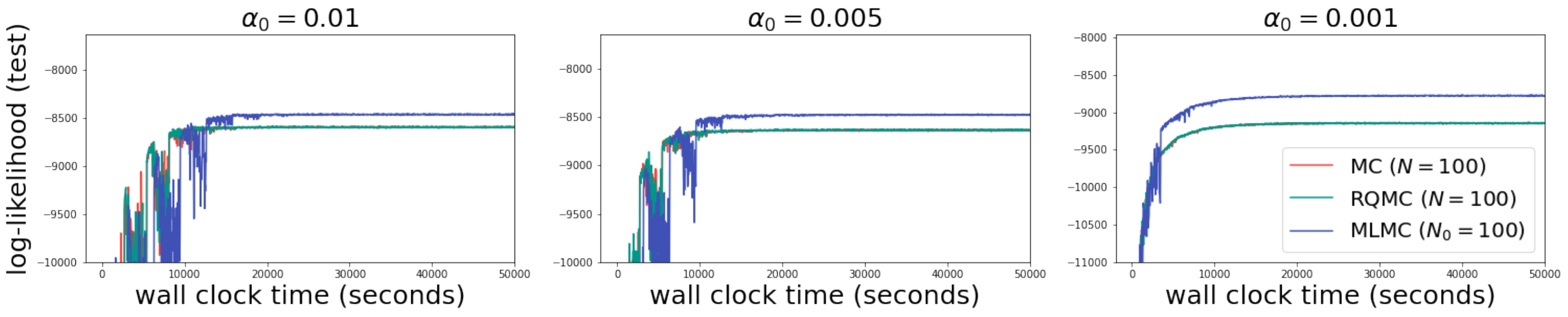}
    \caption{Test log-likelihood (higher is better) for Bayesian logistic regression on the Fashion-MNIST dataset when the initial learning rate $\alpha_{0} = \{0.01,0.005,0.001\}$.}
    \label{fig:blr_mnist_all}
\end{figure}

From the results shown {in Figure}~\ref{fig:blr_mnist_all},
we confirmed that our method {provides a better log-likelihood} than existing methods, although the convergence speed is almost comparable.

\begin{figure}[t]
  \centering
  \includegraphics[scale=0.34]{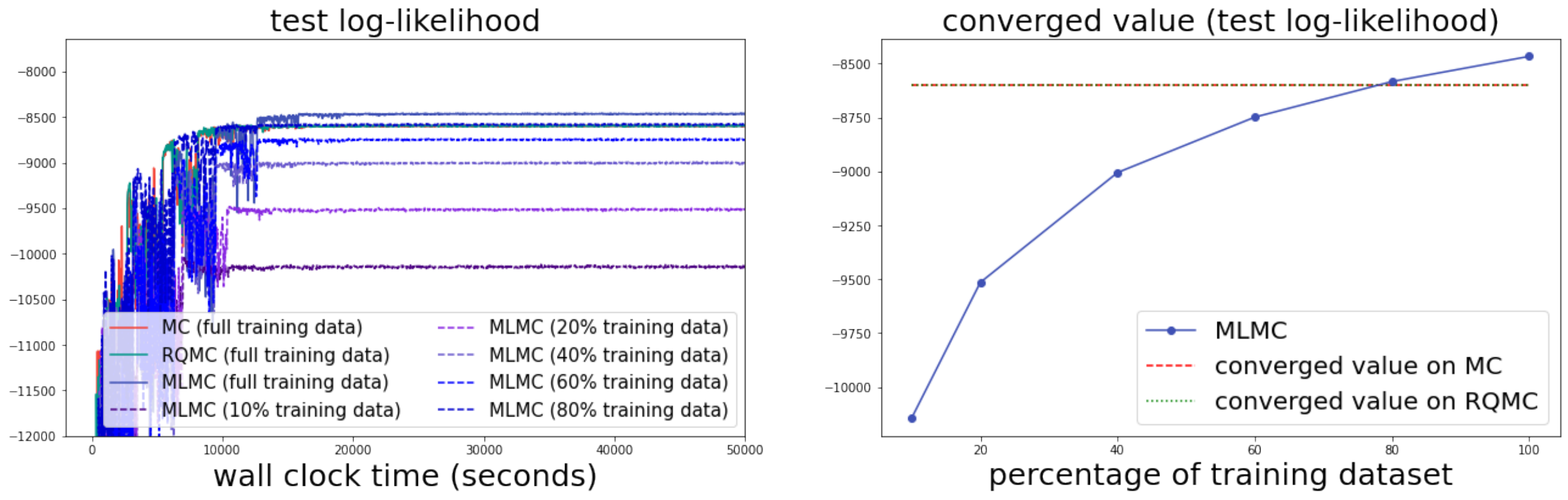}
  \caption{Experimental results when the size of training data changes. Test log-likelihood (higher is better) and its converged value (higher is better) for each percentage are lined up from the left.}
  \label{sample_change}
\end{figure}

\subsubsection{Experimental Results on Data-size Change}
We have confirmed that our method achieves a better test log-likelihood than the baseline method from the experimental results above.
Then, how many training datasets are sufficient for our method to achieve better performance than the baseline method?
{To answer this question}, we performed the following experiments.

Firstly, we adopted Bayesian logistic regression in this experiment and used the same settings as those in the experiments described in Section~\ref{exp:multi}.
Next, we generated five training datasets {with $10\%$, $20\%$, $40\%$, $60\%$, and $80\%$ of the original training data.}
Then, we performed inference on the five reduced training datasets and calculated the log-likelihood for the test data.

The results are shown in Figure~\ref{sample_change}.
From this result, we can see that, even though the training datasets were reduced by about $20\%$, our method achieved almost the same performance as the baseline method performed on the complete training data.
In other words, our method has the potential to perform inference more effectively with fewer data than existing methods.
\section{Conclusion}
\label{conclusion}
We have proposed a new framework for MCVI with the reparameterized gradient estimator, Multilevel Monte Carlo variational inference (MLMCVI).
In the MLMCVI framework, a unique {parameter} update scheme is naturally derived from SGD.
{Moreover}, the sample size for gradient estimation can be adaptively determined according to the gradient variance, which provides the minimum total variance at each optimization step.
Our method can be easily implemented in an automated inference library with an automatic differentiation toolbox, since our method only requires storing the past parameters and using them to estimate the old gradients.

From a theoretical perspective, we {have shown} that the convergence of the weighted average gradient norm is accelerated and that {the learning rate scheduler function can help reduce the degradation of the quality of the MLRG estimator.}
{Furthermore}, three benchmark experiments confirmed that our method achieved comparable or better performance in terms of convergence speed, variance reduction, and SNR than the baseline methods.
We also confirmed that our method provides a higher test log-likelihood in some cases than the baseline methods.

Since our method uses pure MC {sampling}, it can be combined with previously proposed variance reduction methods for the reparameterized gradient based on control variates and importance sampling.
It is also possible to use RQMC sampling by extending the methods in previous studies, such as those by \citet{Giles09b} and \citet{Gerstner13}, to our method.

There are two limitations of our method.
{First}, the theoretical guarantee is shown only when the Gaussian distribution is set as the variational distribution, although commonly used in MCVI; thus, the performance is not guaranteed when other distributions are adopted.
{Second}, in some cases, a truncation is required for $\mathcal{T}(\epsilon;\lambda)$ {from Assumption~\ref{asm:lipshitz2}}, which may cause {biases} in the gradient estimation.
To overcome these limitations, it is necessary to theoretically guarantee the performance of our method {under} more general variational distributions and analyze the bias {induced} by truncation of $\mathcal{T}(\epsilon;\lambda)$.
We leave this part as our future work.

Our idea of recycling past gradients may be useful not only in MCVI but also in the context of stochastic optimization.
Therefore, as our future research, we plan to {extend our method} to a general stochastic optimization algorithm.
\clearpage
\acks{We thank Dr. Ikko Yamane, Dr. Futoshi Futami, and Soma Yokoi for their helpful discussions and comments.
We also thank Dr. Ikko Yamane, Kento Nozawa, Dr. Yoshihiro Nagano, and Han Bao for maintaining the experimental environment.
We also thank Dr. Naonori Ueda, Dr. Tomoharu Iwata, Naoki Marumo, Dr. Masakazu Ishihata, and Dr. Takuma Otsuka of NTT Communication Science Laboratory for their insightful discussions.
We would like to express our deepest gratitude to Professor Masashi Sugiyama for his fruitful advice on our paper.
We are also grateful to many anonymous reviewers who provided insightful reviews at various international conferences over nearly three and a half years before this paper was accepted for publication in JMLR.
We would like to thank the reviewers of JMLR and Dr. Erik Sudderth for their consideration of this paper.
This work was supported by RIKEN Junior Research Associate Program until March, 2021.
M.F. was supported by the University of Tokyo Toyota-Dwango Scholarship for Advanced AI Talents.
M.F. was also supported by JSPS KAKENHI Grant Number 21J11859.}
\appendix
\section{Additional Information on MLMC Method}
\label{app.A}
We introduce additional information on the MLMC method for ease in understanding the background, method, algorithm, and theoretical analysis of our method.

\subsection{Sampling Method and Multilevel Monte Carlo (MLMC)}
When we approximate posterior distributions, Monte Carlo methods are often used for estimating the expectation of
intractable objects with several random samples.
The mean squared error (MSE) of approximation with random samples is a rate of $\mathcal{O}(N^{-1})$,
and an accuracy of $\epsilon$ requires $N = \mathcal{O}(\epsilon^{-2})$ samples.
This rate can be too high for application.
One approach to addressing this high cost is the use of the QMC or RQMC method, in which
the samples are not chosen randomly and independently; instead, they are selected very carefully to reduce
the error~\citep{Giles15}.
In the best cases, the error rate is $\mathcal{O}(N^{-2} \log N^{2d-2})$.
There are many reviews about the QMC approach provided by \citet{Nied92}, \citet{Pierre05}, and \citet{Leobacher14}.

Another approach to improving the computational efficiency is the MLMC method proposed by \citet{Heinrich01}.
This method has been often used in stochastic differential equations for options pricing~\citep{Giles08, Cliffe11, Rhee15}.
In statistics, there are many applications in approximate Bayesian computation~\citep{giles16,jasra17,warne18}.
In a Bayesian framework, \citet{giles16, Giles19} applied MLMC to stochastic gradient MCMC algorithms such as the stochastic gradient Langevin dynamics (SGLD), which discretize the posterior of SDE based on the Multilevel step size and couple them.
        
Because of the linearity of expectations, given a sequence $P_{0},P_{1},\ldots,P_{L-1}$ that approximates $P_{L}$ with increasing accuracy, we have a simple identity:
\begin{equation}
        \label{eq:MLMC}
        \mathbb{E}[P_{L}] = \mathbb{E}[P_{0}] + \sum_{l=1}^{L}\mathbb{E}[P_{l} - P_{l-1}].
        \end{equation}
        We can thus use the following unbiased estimator for $\mathbb{E}[P_{L}]$,
        \begin{equation}
        \label{eq:MLMC-unbiased}
        \begin{split}
                \mathbb{E}[P_{L}] \approx N_{0}^{-1}\sum_{n=1}^{N_{0}}P_{0}^{(0,n)} + \sum_{l=1}^{L} \left\{ N_{l}^{-1} \sum_{n=1}^{N_{l}} (P_{l}^{(l,n)} - P_{l-1}^{(l,n)}) \right\},
        \end{split}
\end{equation}
with the inclusion of $l$ in $(l,n)$, indicating that independent samples are used at each level of correction. 
        
When we define $V_{0},C_{0}$ to be the variance and cost of \emph{one sample} of $P_{0}$, and $\mathbb{V}_{l},C_{l}$ to be the variance and cost of \emph{one sample} of $P_{l} - P_{l-1}$, then, the total variance and cost in Eq.~\ref{eq:MLMC-unbiased} are $\sum_{l=0}^{L}N_{l}^{-1}\mathbb{V}_{l}$ and $\sum_{l=0}^{L}N_{l}C_{l}$, respectively.
        
Thus, if $Y$ is a Multilevel estimator given by
\begin{align*}
         Y = \sum_{l=0}^{L} Y_{l}, \ Y_{l} = N_{l}^{-1} \sum_{n=1}^{N_{l}} (P_{l}^{(l,n)} - P_{l-1}^{(l,n)}),
\end{align*}
with $P_{-1} \equiv 0$, then
\begin{align*}
\mathbb{E}[Y] = \mathbb{E}[P_{L}], \ V[Y] = \sum_{l=0}^{L} N_{l}^{-1} \mathbb{V}_{l}, \ \mathbb{V}_{l} \equiv V[P_{l} - P_{l-1}].
\end{align*}
The MLMC method has been described in detail by \citet{Heinrich01} and \citet{Giles08, Giles15}.

\subsection{Control Variates and Their Relationship with Two-level MLMC}
One of the classic methods to reduce the variance of Monte Carlo samples is control variates~\citep{Glasserman03}.
Let us consider that we want to estimate the expectation of a function $f$, i.e., $\mathbb{E}[f]$.
Then, in control variates, we construct a function $h$ that is highly correlated with $f$ with a known expectation $\mathbb{E}[h]$ and estimate $\mathbb{E}[f]$ via its unbiased estimator from $N$ i.i.d. samples $\omega^{(n)}$ as follows:
\begin{align}
    N^{-1} \sum_{n=1}^{N} \{ f(\omega^{(n)}) - a (h(\omega^{(n)} - \mathbb{E}[h])) \}.
\end{align}
Then, variance is expressed as
\begin{align*}
    V[f(\omega^{(n)})] = V[f(\omega^{(n)})] + a^{2} V[h(\omega^{(n)})] - 2a\textrm{Cov}(f(\omega^{(n)}),h(\omega^{(n)})),
\end{align*}
and the optimal value for $a$ is $\rho \sqrt{V[f]/V[h]}$, 
where $\rho$ is the correlation between $f$ and $h$.
Thus, the variance of this estimator is reduced by a factor of $1 - \rho^{2}$~\citep{Giles08}.

The two-level MLMC method is similar to this method.
According to \citet{Giles13}, if we want to estimate $\mathbb{E}[P_{1}]$ but it is much cheaper to simulate $P_{0}$ that approximates $P_{1}$, then since 
\begin{align}
    \mathbb{E}[P_{1}] = \mathbb{E}[P_{0}] + \mathbb{E}[P_{1} - P_{0}],
\end{align}
we can use the unbiased two-level estimator
\begin{align}
    N_{0}^{-1} \sum_{n=1}^{N_{0}} P_{0}^{(n)} + N_{1}^{-1} \sum_{n=1}^{N_{1}} \bigg( P_{1}^{(n)} - P_{0}^{(n)} \bigg).
\end{align}
There are two different points regarding control variates: the value of $\mathbb{E}[P_{0}]$ is {\it{unknown}} and $a$ takes one.
\section{Proofs}
\label{app.B}
Here, we show the proofs of lemmas and theorems introduced in this paper.

\subsection{Proof of Lemma \ref{lem:MLRG_update}}
\label{lem1:proof}
\begin{proof}
MLRG {at} iteration $T$ can be rewritten as
\begin{align*}
    \nabla_{\lambda_{T}}^{\mathrm{MLRG}} \mathcal{F}(\lambda_{T}) 
    &= \mathbb{E}_{p(\epsilon)} [g_{\lambda_{0}}(\epsilon)] + \sum_{t=1}^{T-1} \bigg( \mathbb{E}_{p(\epsilon)} \bigg[g_{\lambda_{t}}(\epsilon) - g_{\lambda_{t-1}}(\epsilon) \bigg] \bigg) + \mathbb{E}_{p(\epsilon)} \bigg[g_{\lambda_{T}}(\epsilon) - g_{\lambda_{T-1}}(\epsilon)\bigg] \\
    &= \nabla_{\lambda_{T-1}}^{\mathrm{MLRG}} \mathcal{F}(\lambda_{T-1}) + \mathbb{E}_{p(\epsilon)} \bigg[g_{\lambda_{T}}(\epsilon) - g_{\lambda_{T-1}}(\epsilon)\bigg].
\end{align*}
By constructing the unbiased estimator in the above $\nabla_{\lambda_{T}}^{\mathrm{MLRG}} \mathcal{F}(\lambda_{T})$, we obtain
\begin{align*}
    \widehat{\nabla}_{\lambda_{T}}^{\mathrm{MLRG}} \mathcal{F}(\lambda_{T}) = \widehat{\nabla}_{\lambda_{T-1}}^{\mathrm{MLRG}} \mathcal{F}(\lambda_{T-1}) + N_{T}^{-1} \sum_{n=1}^{N_{T}} \bigg[g_{\lambda_{T}}(\epsilon_{(n,T)}) - g_{\lambda_{T-1}}(\epsilon_{(n,T)}) \bigg].
\end{align*}
From the above result, we have the following update rule according to the SGD:
\begin{align*}
    \lambda_{T+1}
    &= \lambda_{T} - \alpha_{T} \bigg(\widehat{\nabla}_{\lambda_{T-1}}^{\mathrm{MLRG}} \mathcal{F}(\lambda_{T-1}) + N_{T}^{-1} \sum_{n=1}^{N_{T}} \bigg[g_{\lambda_{T}}(\epsilon_{(n,T)}) - g_{\lambda_{T-1}}(\epsilon_{(n,T)}) \bigg] \bigg) \\
    &= \lambda_{T} - \frac{\eta_{T}}{\eta_{T-1}} \alpha_{0} \eta_{T-1} \widehat{\nabla}_{\lambda_{T-1}}^{\mathrm{MLRG}} \mathcal{F}(\lambda_{T-1}) - \alpha_{T} N_{T}^{-1} \sum_{n=1}^{N_{T}} \bigg[g_{\lambda_{T}}(\epsilon_{(n,T)}) - g_{\lambda_{T-1}}(\epsilon_{(n,T)}) \bigg] \\
    &= \lambda_{T} + \frac{\eta_{T}}{\eta_{T-1}} (\lambda_{T} - \lambda_{T-1} ) - \alpha_{T} N_{T}^{-1} \sum_{n=1}^{N_{T}} \bigg[g_{\lambda_{T}}(\epsilon_{(n,T)}) - g_{\lambda_{T-1}}(\epsilon_{(n,T)}) \bigg].
\end{align*}
The claim is proved by changing $T$ to $t$.
\end{proof}

\subsection{Proof of Theorem \ref{thm:optimal_sample}}
\label{thm1:proof}
\begin{proof}
According to Assumption~\ref{assmp:comp_cost}, the constrained minimization problem we consider can be expressed as
{
\begin{align*}
\min_{N_{t}} \sum_{t=0}^{T} N_{t}^{-1} \mathbb{V}_{t} \ \ \text{s.t.} \ \ \sum_{t=0}^{T} N_{t} C_{t} \leq M,
\end{align*}
where $M = \sum_{t=0}^{T}C_{t}N_{0}$.
}
For a fixed cost $M$, the variance is minimized by choosing $N_{t}$ to minimize
\begin{align}
\label{optim1}
f(N_{t}) = \sum_{t=0}^{T} N_{t}^{-1} \mathbb{V}_{t} + \mu^{2} \bigg( \sum_{t=0}^{T} N_{t}C_{t} - M \bigg),
\end{align}
for some value of the Lagrange multiplier $\mu^{2} \ (\mu > 0)$. 
\diff{
By solving the above with respect to $N_{t}$, we obtain
\begin{align}
\label{optim-sample1}
N_{t} = \frac{1}{\mu} \sqrt{ \frac{\mathbb{V}_{t}}{C_{t}} } \ 
(\because \mu > 0, C_{t} > 0, \mathbb{V}_{t} > 0).
\end{align}
By substituting \ref{optim-sample1} for \ref{optim1}, we obtain
\begin{align}
\label{optim2}
f(N_{t}) = 2 \mu \sum_{t=0}^{T} \sqrt{\mathbb{V}_{t}C_{t}} - \mu^{2} M.
\end{align}
By differentiating $f(N_{t})$ in \ref{optim2} and solving it with respect to $\mu$, we obtain
\begin{align}
\label{optimal-mu}
\mu = \frac{1}{M} \sum_{t=0}^{T} \sqrt{\mathbb{V}_{t}C_{t}}.
\end{align}
We substitute (\ref{optimal-mu}) for (\ref{optim-sample1}):
\begin{align}
\label{eq:optimal_n}
N_{t} = \frac{M}{\sum_{t=0}^{T} \sqrt{\mathbb{V}_{t} C_{t}}} \sqrt{ \frac{\mathbb{V}_{t}}{C_{t}} }.
\end{align}
By considering the ratio of $N_{t+1}$ to $N_{t}$ \diff{for $t \geq 1$ and according to Assumption~\ref{assmp:comp_cost}}, we obtain
\begin{align*}
\frac{N_{t+1}}{N_{t}} &= \frac{M}{\sum_{t=0}^{T} \sqrt{\mathbb{V}_{t}C_{t}}} \sqrt{ \frac{\mathbb{V}_{t+1}}{C_{t+1}}}
\cdot \frac{\sum_{t=0}^{T} \sqrt{\mathbb{V}_{t}C_{t}}}{M} \sqrt{ \frac{C_{t}}{\mathbb{V}_{t}}} \\
&= \sqrt{\frac{\mathbb{V}_{t+1}}{2\nu d}} \cdot \sqrt{ \frac{2\nu d}{\mathbb{V}_{t}}} \\
&= \sqrt{\frac{\mathbb{V}_{t+1}}{\mathbb{V}_{t}}}.
\end{align*}
According to this result, the optimal sample size
$N_{t}$ can be expressed as
\begin{align*}
N_{t+1} &= \sqrt{\frac{\mathbb{V}_{t+1}}{\mathbb{V}_{t}}} N_{t},
\end{align*}
for $t=1,\ldots,T$.
Furthermore, from Eq.~\ref{eq:optimal_n}, we can see that when $t=1$,
\begin{align*}
    N_{1} &= \frac{M}{\sum_{t=0}^{T} \sqrt{\mathbb{V}_{t} C_{t}}} \sqrt{ \frac{\mathbb{V}_{1}}{C_{1}} } \\
    &= \frac{\nu d N_{0} + 2\nu d N_{0}T}{\sqrt{\nu d \mathbb{V}_{0}} + \sqrt{2\nu d} \sum_{t=1}^{T} \sqrt{\mathbb{V}_{t}}} \sqrt{ \frac{\mathbb{V}_{1}}{2\nu d} } \\
    &= \frac{(2T+1)\sqrt{\mathbb{V}_{1}}}{\sqrt{2\mathbb{V}_{0}} + 2 \sum_{t=1}^{T} \sqrt{\mathbb{V}_{t}}}N_{0}.
\end{align*}
Thus, the claim is proved.}
\end{proof}

\subsection{Proof of Proposition \ref{prop:one-sample_variance}}
\label{prop1:proof}
\begin{proof}
According to Assumption \ref{asm:lipshitz2}, we obtain the Lipschitz condition given by
\begin{align*}
    ||g_{\lambda_{t}}(\epsilon) - g_{\lambda_{t-1}}(\epsilon) ||_{2}^{2} \leq K_{1} ||\mathcal{T}(\epsilon; \lambda_{t}
        ) - \mathcal{T}(\epsilon; \lambda_{t-1}
        )||_{2}^{2} < \infty.
\end{align*}
According to Assumption~\ref{asm:gaussian}, all of the variational parameters can be expressed as a single vector $\lambda = (\bf{m,v})$ and its transformation $\mathcal{T}(\epsilon; \lambda_{t})$ can be written as $\mathcal{T}(\epsilon; \lambda_{t}) = {\bf{m}}_{t} + {\bf{v}}_{t} \cdot \epsilon$.
From these assumptions, we obtain
\begin{align}
\label{eq:deff_bound}
    ||g_{\lambda_{t}}(\epsilon) - g_{\lambda_{t-1}}(\epsilon) ||_{2}^{2} 
    &\leq K_{1} ||\mathcal{T}(\epsilon; \lambda_{t}
        ) - \mathcal{T}(\epsilon; \lambda_{t-1}
        )||_{2}^{2} \notag \\
        &= K_{1} \| ({\bf{m}}_{t} - {\bf{m}}_{t-1}) + ({\bf{v}}_{t} - {\bf{v}}_{t-1}) \cdot \epsilon \|_{2}^{2} \notag \\
        &\leq K_{1} (\|{\bf{m}}_{t} - {\bf{m}}_{t-1} \|_{2}^{2} + \|({\bf{v}}_{t} - {\bf{v}}_{t-1}) \cdot \epsilon \|_{2}^{2})
        \notag \\
        &\leq K_{1} (\|{\bf{m}}_{t} - {\bf{m}}_{t-1} \|_{2}^{2} + \|{\bf{v}}_{t} - {\bf{v}}_{t-1}\|_{2}^{2} \cdot \| \epsilon \|_{2}^{2}).
\end{align}
The third and fourth lines are obtained using the triangle inequality and Cauchy--Schwarz inequality, respectively.
Because of the MLRG update rule with decreasing learning rate from Lemma~\ref{lem:MLRG_update},
we obtain
\begin{align*}
    \|{\bf{m}}_{t} &- {\bf{m}}_{t-1} \|_{2}^{2}
    = \bigg\|\frac{\eta_{t-1}}{\eta_{t-2}} ({\bf{m}}_{t-1} - {\bf{m}}_{t-2}) - \alpha_{t-1}A_{t-1} \bigg\|_{2}^{2} \\
    &= \bigg\|\frac{\eta_{t-1}}{\underset{=1}{\underline{\eta_{0}}}} ({\bf{m}}_{1} - {\bf{m}}_{0}) - \frac{\eta_{t-1}}{\underset{=1}{\underline{\eta_{0}}}}\cdot \alpha_{1} A_{1} - \frac{\eta_{t-1}}{\eta_{1}}\cdot \alpha_{2} A_{2} - \cdots - \frac{\eta_{t-1}}{\eta_{t-3}}\cdot \alpha_{t-2} A_{t-2} - \alpha_{t-1} A_{t-1} \bigg\|_{2}^{2} \\
    &= \eta_{t-1}^{2}\bigg\| - \alpha_{0}\hat{g}_{{\bf m}_{0}}(\epsilon_{1:N_{0}}) - \alpha_{1} A_{1} - \alpha_{0} \cdot \frac{\eta_{2}}{\eta_{1}} A_{2} - \cdots - \alpha_{0} \cdot \frac{\eta_{t-2}}{\eta_{t-3}}  A_{t-2} - \alpha_{0} A_{t-1} \bigg\|_{2}^{2} \\
    &= \alpha_{0}^{2}\eta_{t-1}^{2}\bigg\| \hat{g}_{{\bf m}_{0}}(\epsilon_{1:N_{0}}) + \eta_{1} A_{1} + \frac{\eta_{2}}{\eta_{1}} A_{2} + \cdots + \frac{\eta_{t-2}}{\eta_{t-3}}  A_{t-2} + A_{t-1} \bigg\|_{2}^{2}.
\end{align*}
{At} iteration $t$, the term $S = \hat{g}_{{\bf m}_{0}}(\epsilon_{1:N_{0}}) + \eta_{1} A_{1} + \frac{\eta_{2}}{\eta_{1}} A_{2} + \cdots + \frac{\eta_{t-2}}{\eta_{t-3}}  A_{t-2}$
is constant.
By using the triangle inequality, we obtain
\begin{align*}
    \|{\bf{m}}_{t} - {\bf{m}}_{t-1} \|_{2}^{2} = \alpha_{0}^{2}\eta_{t-1}^{2}\|S + A_{t-1} \|_{2}^{2}
    \leq \alpha_{0}^{2}\eta_{t-1}^{2} (\|S\|_{2}^{2} + \|A_{t-1} \|_{2}^{2}).
\end{align*}
By taking the expectation over $p(\epsilon)$, we obtain
\begin{align*}
    \mathbb{E}_{p(\epsilon)}[\|{\bf{m}}_{t} - {\bf{m}}_{t-1} \|_{2}^{2}] \leq \alpha_{0}^{2}\eta_{t-1}^{2} (\|S\|_{2}^{2} + \mathbb{E}_{p(\epsilon)}[\|A_{t-1} \|_{2}^{2}]).
\end{align*}
Here,
\begin{align*}
    \mathbb{E}_{p(\epsilon)}[\|A_{t-1} \|_{2}^{2}] &= \mathbb{E}_{p(\epsilon)}\bigg[\bigg\|N_{t-1}^{-1} \sum_{n=1}^{N_{t-1}} \bigg[g_{\lambda_{t-1}}(\epsilon_{(n,t-1)}) - g_{\lambda_{t-2}}(\epsilon_{(n,t-1)}) \bigg]\bigg\|_{2}^{2}\bigg] \\
    &\leq \mathbb{E}_{p(\epsilon)}\bigg[\| g_{\lambda_{t-1}}(\epsilon_{(1,t-1)}) - g_{\lambda_{t-2}}(\epsilon_{(1,t-1)}) \|_{2}^{2}\bigg] \\ 
    &\leq K_{1}\mathbb{E}_{p(\epsilon)}\bigg[ \|\mathcal{T}(\epsilon_{(1,t-1)}; \lambda_{t-1}
        ) - \mathcal{T}(\epsilon_{(1,t-1)}; \lambda_{t-2}
        )\|_{2}^{2}\bigg]  < \infty \ (\textrm{Assumption \ref{asm:lipshitz2}}),
\end{align*}
where we assume that $\max \{ g_{\lambda_{t-1}}(\epsilon_{(n,t-1)}) - g_{\lambda_{t-2}}(\epsilon_{(n,t-1)}) |1 \leq n \leq N_{t-1}\} = g_{\lambda_{t-1}}(\epsilon_{(1,t-1)}) - g_{\lambda_{t-2}}(\epsilon_{(1,t-1)})$.

From these results, the magnitude of $\mathbb{E}_{p(\epsilon)}[\|{\bf{m}}_{t} - {\bf{m}}_{t-1} \|_{2}^{2}]$ can be seen as $\mathcal{O}(\eta_{t-1}^{2})$.
The same results are obtained for the term $\mathbb{E}_{p(\epsilon)}[\|{\bf{v}}_{t} - {\bf{v}}_{t-1} \|_{2}^{2}]$.

Since $\epsilon \overset{\mathrm{i.i.d.}}{\sim} p(\epsilon) \in \mathbb{R}^{d}$, the expectation of $\| \epsilon \|_{2}^{2}$ is obtained as
\begin{align*}
    \mathbb{E}_{p(\epsilon)}\bigg[\| \epsilon \|_{2}^{2}\bigg] 
    &= \mathbb{E}_{p(\epsilon)}[\epsilon_{(1)}^{2}] +  \mathbb{E}_{p(\epsilon)}[\epsilon_{(2)}^{2}] + \cdots +
    \mathbb{E}_{p(\epsilon)}[\epsilon_{(d)}^{2}]
    = d\mathbb{E}_{p(\epsilon)}[\epsilon_{(1)}^{2}].
\end{align*}
If we consider $\mathbb{E}_{p(\epsilon)}[\epsilon_{(1)}^{2}] \leq \delta (\geq 0)$, $\mathbb{E}_{p(\epsilon)}[\|{\bf{m}}_{t} - {\bf{m}}_{t-1} \|_{2}^{2}] \leq c_{1}\eta_{t-1}^{2}$, and $\mathbb{E}_{p(\epsilon)}[\|{\bf{v}}_{t} - {\bf{v}}_{t-1} \|_{2}^{2}] \leq c_{2}\eta_{t-1}^{2}$, we can obtain
\begin{align*}
    \mathbb{E}_{p(\epsilon)}[\|g_{\lambda_{t}}(\epsilon) - g_{\lambda_{t-1}}(\epsilon) \|_{2}^{2}] &\leq K_{1}(c_{1}\eta_{t-1}^{2} + d\delta c_{2}\eta_{t-1}^{2})
    = \eta_{t-1}^{2} K_{1} (c_{1} + d \delta c_{2}),
\end{align*}
where $\delta$, $C_{1}$, and $C_{2}$ are positive constants.

As $t \rightarrow \infty$, we can see that $\mathbb{E}_{p(\epsilon)} [||g_{\lambda_{t}}(\epsilon) - g_{\lambda_{t-1}}(\epsilon) ||_{2}^{2}] = \mathcal{O}(\eta_{t-1}^{2})$.
Furthermore, $\mathbb{V}_{t}$ is typically similar in magnitude to $\mathbb{E}_{p(\epsilon)} [||g_{\lambda_{t}}(\epsilon) - g_{\lambda_{t-1}}(\epsilon) ||_{2}^{2}]$~\citep{Giles15} because
\begin{align*}
    \mathbb{V}[g_{\lambda_{t}}(\epsilon) - g_{\lambda_{t-1}}(\epsilon)]
    &= \mathbb{E}_{p(\epsilon)} \bigg[\left\|g_{\lambda_{t}}(\epsilon) - g_{\lambda_{t-1}}(\epsilon) \right\|_{2}^{2} \bigg] -  \left\| \mathbb{E}_{p(\epsilon)} \bigg[ g_{\lambda_{t}}(\epsilon) - g_{\lambda_{t-1}}(\epsilon) \bigg] \right\|_{2}^{2} \\
    &\leq \mathbb{E}_{p(\epsilon)} \bigg[\left\|g_{\lambda_{t}}(\epsilon) - g_{\lambda_{t-1}}(\epsilon) \right\|_{2}^{2} \bigg].
\end{align*}
Therefore, we find that $\mathbb{V}[g_{\lambda_{t}}(\epsilon) - g_{\lambda_{t-1}}(\epsilon)] = \mathbb{V}_{t} =  \mathcal{O}(\eta_{t-1}^{2})$.

Considering the fact that $\alpha_{t} = \alpha_{0}\eta_{t} \overset{t \rightarrow \infty}{\rightarrow} 0$, the one-sample variance $\mathbb{V}_{t}$ becomes $0$ asymptotically as iteration proceeds.
Thus, the claim is proved.
\end{proof}

\diff{
\subsection{Proof of Theorem~\ref{thm:alt_optimal_sample}}
\label{thm2:proof}
\begin{proof}
Since the order of variance in gradient estimation via Monte Carlo estimation is $\mathcal{O}(N^{-1})$, we can see the order of one-sample gradient variance at iteration $t = 0$ as $\mathbb{V}_{0} = \mathcal{O}(1)$.
In addition, from Proposition~\ref{prop:one-sample_variance}, the order of one-sample gradient variance for $t \geq 1$ is $\mathcal{O}(\eta_{t-1}^{2})$.
Then, we have the upper bound of the total variance as 
\begin{align*}
    \sum_{t=0}^{T}N_{t}^{-1} \mathbb{V}_{t}
    \leq N_{0}^{-1} \kappa + \sum_{t=1}^{T} N_{t}^{-1} \kappa \eta_{t-1}^{2},
\end{align*}
where $\kappa$ is a positive constant.
As Theorem~\ref{thm:optimal_sample}, by considering the constraint optimization problem based on the above upper bound, we have the following optimization problem:
\begin{align*}
\min_{N_{t}} N_{0}^{-1} \kappa + \sum_{t=1}^{T} N_{t}^{-1} \kappa \eta_{t-1}^{2} \ \ \text{s.t.} \ \ \sum_{t=0}^{T} N_{t} C_{t} \leq M,
\end{align*}
where $M = \sum_{t=0}^{T} N_{0} C_{t}$.
By solving the above with respect to $N_{t}$ in the same way as Theorem~\ref{thm:optimal_sample}, we obtain
$N_{t} = \frac{M}{\sqrt{\nu d\kappa} + \sqrt{2\nu d\kappa} \sum_{t=1}^{T}\eta_{t-1}}\sqrt{\frac{\kappa \eta_{t-1}^{2}}{C_{t}}} = \frac{M}{\sqrt{\nu d} + \sqrt{2\nu d} \sum_{t=1}^{T}\eta_{t-1}}\sqrt{\frac{ \eta_{t-1}^{2}}{C_{t}}}$ and
$N_{t+1} = \frac{\eta_{t}}{\eta_{t-1}}N_{t}$ for $t \geq 1$.
Since $\eta_{0} = 1$, we obtain
\begin{align*}
    N_{t+1} 
    = \frac{\eta_{t}}{\eta_{t-1}} \cdot \frac{\eta_{t-1}}{\eta_{t-2}}N_{t-1} 
    &= \frac{\eta_{t}}{\eta_{t-1}} \cdot \frac{\eta_{t-1}}{\eta_{t-2}} \cdots
    \frac{\eta_{1}}{\eta_{0}} N_{1} \\
    &= \eta_{t}N_{1}.
\end{align*}
Furthermore, when $t=1$, we have
\begin{align*}
    N_{1} 
    = \frac{M}{\sqrt{\nu d} + \sqrt{2\nu d} \sum_{t=1}^{T}\eta_{t-1}}\sqrt{\frac{\mathbb\eta_{0}^{2}}{C_{1}}}
    &= \frac{\nu d N_{0} + 2\nu d N_{0}T}{\sqrt{\nu d} + \sqrt{2\nu d} \sum_{t=1}^{T}\eta_{t-1}}
    \sqrt{\frac{1}{2\nu d}} \\
    &= \frac{2T + 1}{\sqrt{2} + 2 \sum_{t=1}^{T}\eta_{t-1}}N_{0}.
\end{align*}
According to the constraint condition, $N_{1}$ should satisfy
\begin{align*}
    N_{0}C_{0} + N_{1}C_{1} \leq N_{0}C_{0} + N_{0}C_{1}
    \implies N_{1} \leq N_{0}.
\end{align*}
In addition, since $\eta_{t-1} \leq 1$, we have
\begin{align*}
  \frac{2T + 1}{\sqrt{2} + 2 \sum_{t=1}^{T}\eta_{t-1}}N_{0}
  \geq \frac{2T + 1}{2T + \sqrt{2}}N_{0}.
\end{align*}
From these, we have the following relationship:
\begin{align*}
    \frac{2T + 1}{2T + \sqrt{2}}N_{0} \leq N_{1} \leq N_{0}.
\end{align*}
Thus, the claim is proved.
\end{proof}
}

\subsection{Proof of Lemma~\ref{cor:optimal}}
\label{cor1:proof}
\begin{proof}
\diff{From Definition~\ref{def:low_total_cost_condition}, since $r_{0} = 0$, we can see
\begin{align*}
    \sum_{t=0}^{T} (N_{t}+r_{t})C_{t}
    = \nu d N_{0} + 2\nu d \sum_{t=1}^{T} (N_{t}+r_{t}).
\end{align*}
Then, we obtain
\begin{align*}
    \sum_{t=0}^{T} N_{t}C_{t} - \sum_{t=0}^{T} (N_{t}+r_{t})C_{t}
    &= \nu d N_{0} + 2\nu d N_{0}T - \nu d N_{0} - 2\nu d \sum_{t=1}^{T} (N_{t}+r_{t}) \\
    &\propto N_{0}T - \sum_{t=1}^{T} (N_{t}+r_{t}) \\
    &= N_{0}T - \sum_{t=1}^{T} N_{t} - \sum_{t=1}^{T} r_{t} \\
    &> N_{0}T - \sum_{t=1}^{T} N_{t} - T \\
    &= N_{0}T - T - N_{1}\sum_{t=1}^{T} \eta_{t} \\
    &= N_{0}T - T - N_{0}\sum_{t=1}^{T} \eta_{t}.
\end{align*}
By setting the learning rate scheduler function $\eta_{t}$ as $\eta_{t} = 1$ for $1 \leq t \leq l$, where $l \in \mathbb{Z}$ and $1 \leq l \leq T-1$, we can rearrange the above as
\begin{align*}
    N_{0}T - T - N_{0}\sum_{t=1}^{T} \eta_{t}
    = N_{0}T - T - N_{0}l - N_{0}\sum_{t=l+1}^{T} \eta_{t}.
\end{align*}
Then, if we set $\eta_{t} \leq 1 - \frac{2}{N_{0}}$ for $l+1 \leq t \leq T$, we obtain
\begin{align*}
    N_{0}T - T - N_{0}l - N_{0}\sum_{t=l+1}^{T} \eta_{t}
    &\geq N_{0}T - T - N_{0}l - N_{0}(T-l) \bigg( 1 - \frac{2}{N_{0}}\bigg) \\
    &= N_{0}T - T - N_{0}l - N_{0}(T-l) + 2(T-l) \\
    &= T - 2l.
\end{align*}
By assuming $T \geq 2$ and taking $l$ as $1 \leq l \leq \lfloor \frac{T}{2} \rfloor$, where $\lfloor x \rfloor = \max \{k \in \mathbb{Z}| k \leq x \}$, we obtain
\begin{align*}
    T - 2l
    &\geq T - 2\bigg\lfloor \frac{T}{2} \bigg\rfloor \\
    &\geq T - 2 \cdot \frac{T}{2}
    = 0.
\end{align*}
Thus, 
\begin{align*}
    \sum_{t=0}^{T} N_{t}C_{t} - \sum_{t=0}^{T} (N_{t}+r_{t})C_{t} > 0,
\end{align*}
and the claim is proved.}

\end{proof}

\subsection{Proof of Lemma \ref{lem:var_order}}
\label{lem3:proof}
\begin{proof}
According to Proposition \ref{prop:one-sample_variance}, the order of $\mathbb{V}_{t}$ is $\mathcal{O}(\eta_{t-1}^{2})$, i.e., $\mathbb{V}_{t} \leq \kappa_{t}\eta_{t-1}^{2},$ where $\kappa_{t}$ is a positive constant at the iteration $t$.
Then, we obtain the following inequality:
\begin{align*}
    \mathbb{V}[\widehat{\nabla}_{\lambda_{t}}^{\mathrm{MLRG}} \mathcal{F}(\lambda_{t})] &= \alpha_{t}^{-2} \mathbb{V}[ \alpha_{t} \widehat{\nabla}_{\lambda_{t}} \mathcal{F}(\lambda_{t})] \\
    &= \alpha_{t}^{-2} \mathbb{V}\bigg[ \frac{\eta_{t}}{\eta_{t-1}} (\lambda_{t} - \lambda_{t-1} ) - \alpha_{t} N_{t}^{-1} \sum_{n=1}^{N_{t}} \bigg[g_{\lambda_{t}}(\epsilon_{(n,t)}) - g_{\lambda_{t-1}}(\epsilon_{(n,t)}) \bigg] \bigg] \\
    &= \mathbb{V}\bigg[ N_{t}^{-1} \sum_{n=1}^{N_{t}} \bigg[g_{\lambda_{t}}(\epsilon_{(n,t)}) - g_{\lambda_{t-1}}(\epsilon_{(n,t)}) \bigg] \bigg] \\
    &= N_{t}^{-1} \mathbb{V}\bigg[ g_{\lambda_{t}}(\epsilon_{(1,t)}) - g_{\lambda_{t-1}}(\epsilon_{(1,t)})  \bigg] 
    \leq N_{t}^{-1} \kappa_{t} \eta_{t-1}^{2}.
\end{align*}
Therefore,
\begin{align*}
    \mathbb{V}[\widehat{\nabla}_{\lambda_{t}}^{\mathrm{MLRG}} \mathcal{F}(\lambda_{t})] \leq N_{t}^{-1} \kappa_{t} \eta_{t-1}^{2}
    = \kappa_{t} \cdot \eta_{t-1}^{2} N_{t}^{-1}
    = \mathcal{O}(\eta_{t-1}^{2} N_{t}^{-1}),
\end{align*}
and the claim is proved.
\end{proof}

\subsection{Proof of Theorem \ref{thm:norm_MC,RQMC}}
\label{thm3:proof}
\begin{proof}
Accorging to Assumption \ref{asm:lipschitz}, we have $\mathcal{F}(\lambda) \leq
        \mathcal{F}(\bar{\lambda}) + \nabla_{\bar{\lambda}} \mathcal{F}(\bar{\lambda})^{\top} (\lambda - \bar{\lambda})
        + \frac{1}{2} K_{2} ||\lambda - \bar{\lambda}||_{2}^{2}, \forall \lambda,\bar{\lambda}$.
        By using the SGD update rule,
        we obtain $\lambda_{t+1} - \lambda_{t} = - \alpha_{t} \hat{g}_{\lambda_{t}}(\epsilon_{1:N})$. 
        Thus, when we set $\lambda = \lambda_{t+1}$ and $\bar{\lambda} = \lambda_{t}$, this assumption can be expressed as
        \begin{align*}
            \mathcal{F}(\lambda_{t+1}) - \mathcal{F}(\lambda_{t})
            &\leq \nabla_{\lambda_{t}} \mathcal{F}(\lambda_{t})^{\top} (\lambda_{t+1} - \lambda_{t})
            + \frac{1}{2}K_{2} ||\lambda_{t+1} - \lambda_{t}||_{2}^{2} \\
            &= - \alpha_{t} \nabla_{\lambda_{t}} \mathcal{F}(\lambda_{t})^{\top} \hat{g}_{\lambda_{t}}(\epsilon_{1:N_{t}})
            + \frac{\alpha_{t}^{2}K_{2}}{2} ||\hat{g}_{\lambda_{t}}(\epsilon_{1:N})||_{2}^{2}.
        \end{align*}
        By taking expectation in the above with respect to $\epsilon_{1:N} \sim p(\epsilon)$, we obtain
        \begin{align*}
            \mathbb{E}_{p(\epsilon_{1:N})}[\mathcal{F}(\lambda_{t+1}) - \mathcal{F}(\lambda_{t})]
            &\leq - \alpha_{t} \nabla_{\lambda_{t}} \mathcal{F}(\lambda_{t})^{\top} \mathbb{E}_{p(\epsilon_{1:N})}[\hat{g}_{\lambda_{t}}(\epsilon_{1:N})]
            + \frac{\alpha_{t}^{2}K_{2}}{2} \mathbb{E}_{p(\epsilon_{1:N})}\bigg[ \left\| \hat{g}_{\lambda_{t}}(\epsilon_{1:N}) \right\|_{2}^{2} \bigg].
        \end{align*}
        Since $\mathbb{E}_{p(\epsilon_{1:N})}[\hat{g}_{\lambda_{t}}(\epsilon_{1:N})] = \nabla_{\lambda_{t}}\mathcal{F}(\lambda_{t})$ and $\mathbb{E}_{p(\epsilon_{1:N})}[ ||\hat{g}_{\lambda_{t}}(\epsilon_{1:N})||_{2}^{2} ] = 
        \mathbb{V}[\hat{g}_{\lambda_{t}}(\epsilon_{1:N})] + ||\mathbb{E}_{p(\epsilon_{1:N})} [\hat{g}_{\lambda_{t}}(\epsilon_{1:N})]||_{2}^{2}$, we obtain
        \begin{align*}
            \mathbb{E}_{p(\epsilon_{1:N})}&[\mathcal{F}(\lambda_{t+1}) - \mathcal{F}(\lambda_{t})] \\
            &\leq - \alpha_{t} ||\nabla_{\lambda_{t}} \mathcal{F}(\lambda_{t})||_{2}^{2} + \frac{\alpha_{t}^{2} K_{2}}{2}
            \bigg(\mathbb{V}[\hat{g}_{\lambda_{t}}(\epsilon_{1:N})] + \left\| \mathbb{E}_{p(\epsilon_{1:N})} [\hat{g}_{\lambda_{t}}(\epsilon_{1:N})] \right\|_{2}^{2} \bigg).
        \end{align*}
        Again, since $\mathbb{E}_{p(\epsilon_{1:N})} [\hat{g}_{\lambda_{t}}(\epsilon_{1:N})]
        = \nabla_{\lambda_{t}} \mathcal{F}(\lambda_{t})$, we can rewrite the above equation as
        \begin{align*}
            \mathbb{E}_{p(\epsilon_{1:N})}[\mathcal{F}(\lambda_{t+1}) - \mathcal{F}(\lambda_{t})] &\leq \frac{\alpha_{t}^{2}K_{2}}{2} \mathbb{V}[\hat{g}_{\lambda_{t}}(\epsilon_{1:N})] + 
            \bigg(\frac{\alpha_{t}^{2}K_{2}}{2} - \alpha_{t} \bigg) ||\nabla_{\lambda_{t}} \mathcal{F}(\lambda_{t})||_{2}^{2}.
        \end{align*}
        By summing $t=1,2,\dots,T$ and taking the total expectation, we obtain
        \begin{align*}
            \mathbb{E}[\mathcal{F}(\lambda_{T}) - \mathcal{F}(\lambda_{1})]
            &\leq \frac{K_{2}}{2} \sum_{t=1}^{T}\alpha_{t}^{2} \mathbb{E}\bigg[\mathbb{V}[\hat{g}_{\lambda_{t}}(\epsilon_{1:N})]\bigg]
            +
            \sum_{t=1}^{T}\bigg(\frac{\alpha_{t}^{2}K_{2}}{2} - \alpha_{t} \bigg) 
             \mathbb{E}\bigg[||\nabla_{\lambda_{t}} \mathcal{F}(\lambda_{t})||_{2}^{2} \bigg].
        \end{align*}
        Since $\mathcal{F}(\lambda^{*}) - \mathcal{F}(\lambda_{1}) \leq \mathbb{E}[\mathcal{F}(\lambda_{T}) - \mathcal{F}(\lambda_{1})]$, 
        where $\lambda_{1}$ is deterministic and $\lambda^{*}$ is the optimal parameter, we obtain the following inequality by dividing the inequality by $A_{T} = \sum_{t=1}^{T} \alpha_{t}$:
        \begin{align*}
            \frac{1}{A_{T}} [\mathcal{F}(\lambda^{*}) - \mathcal{F}(\lambda_{1})] &\leq 
            \frac{K_{2}}{2A_{T}} \sum_{t=1}^{T}\alpha_{t}^{2} \mathbb{E}\bigg[\mathbb{V}[\hat{g}_{\lambda_{t}}(\epsilon_{1:N})]\bigg]
            + \frac{1}{A_{T}}\sum_{t=1}^{T}\bigg(\frac{\alpha_{t}^{2}K_{2}}{2} - \alpha_{t} \bigg) 
             \mathbb{E}\bigg[||\nabla_{\lambda_{t}} \mathcal{F}(\lambda_{t})||_{2}^{2} \bigg].
        \end{align*}
        Therefore, we can obtain
        \begin{align*}
            \frac{1}{A_{T}}\sum_{t=1}^{T} &\alpha_{t}\mathbb{E}\bigg[||\nabla_{\lambda_{t}} \mathcal{F}(\lambda_{t})||_{2}^{2}\bigg] \\ 
            &\leq \frac{1}{A_{T}} [\mathcal{F}(\lambda_{1}) - \mathcal{F}(\lambda^{*})] + \frac{K_{2}}{2A_{T}}\sum_{t=1}^{T} \alpha_{t}^{2} \mathbb{E}\bigg[||\nabla_{\lambda_{t}} \mathcal{F}(\lambda_{t})||_{2}^{2} \bigg]
            + \frac{K_{2}}{2A_{T}} \sum_{t=1}^{T}\alpha_{t}^{2} \mathbb{E}\bigg[\mathbb{V}[\hat{g}_{\lambda_{t}}(\epsilon_{1:N})]\bigg].
        \end{align*}
        If we estimate the gradient values using MC samples, we can see that $\mathbb{V}[\hat{g}_{\lambda_{t}}(\epsilon_{1:N})] \leq \kappa N^{-1}$ for some positive constant $\kappa$; therefore, we obtain
        
        \begin{align*}
            \frac{1}{A_{T}}\sum_{t=1}^{T} &\alpha_{t}\mathbb{E}\bigg[||\nabla_{\lambda_{t}} \mathcal{F}(\lambda_{t})||_{2}^{2} \bigg] \\
            &\leq \frac{1}{A_{T}} [\mathcal{F}(\lambda_{1}) - \mathcal{F}(\lambda^{*})] + \frac{K_{2}}{2A_{T}}\sum_{t=1}^{T} \alpha_{t}^{2} \mathbb{E}\bigg[||\nabla_{\lambda_{t}} \mathcal{F}(\lambda_{t})||_{2}^{2}\bigg]
            + \frac{K_{2}}{2A_{T}} \sum_{t=1}^{T}\alpha_{t}^{2} \mathbb{E}\bigg[\mathbb{V}[\hat{g}_{\lambda_{t}}(\epsilon_{1:N})]\bigg] \\
            &\leq \frac{1}{A_{T}} [\mathcal{F}(\lambda_{1}) - \mathcal{F}(\lambda^{*})] + \frac{K_{2}}{2A_{T}}\sum_{t=1}^{T} \alpha_{t}^{2} \mathbb{E}\bigg[||\nabla_{\lambda_{t}} \mathcal{F}(\lambda_{t})||_{2}^{2}\bigg]
            + \frac{\kappa \cdot K_{2}}{2A_{T}N} \sum_{t=1}^{T}\alpha_{t}^{2} \\
            &= \frac{1}{A_{T}} [\mathcal{F}(\lambda_{1}) - \mathcal{F}(\lambda^{*})] + \frac{\alpha_{0}^{2}K_{2}}{2A_{T}}\sum_{t=1}^{T}  \eta_{t}^{2} \bigg( \mathbb{E}\bigg[||\nabla_{\lambda_{t}} \mathcal{F}(\lambda_{t})||_{2}^{2}\bigg]
            + \frac{\kappa}{N} \bigg).
        \end{align*}
        The last term is derived from $\alpha_{t} = \alpha_{0} \cdot \eta_{t}$.
        When estimating the gradient value using RQMC samples, since the order of variance is $\mathcal{O}(N^{-2})$, we can obtain the same as above as follows:
        \begin{align*}
            \frac{1}{A_{T}}\sum_{t=1}^{T} \alpha_{t}\mathbb{E}\bigg[||\nabla_{\lambda_{t}} \mathcal{F}(\lambda_{t})||_{2}^{2}\bigg] &\leq \frac{1}{A_{T}} [\mathcal{F}(\lambda_{1}) - \mathcal{F}(\lambda^{*})] + \frac{\alpha_{0}^{2}K_{2}}{2A_{T}}\sum_{t=1}^{T}  \eta_{t}^{2} \bigg( \mathbb{E}\bigg[||\nabla_{\lambda_{t}} \mathcal{F}(\lambda_{t})||_{2}^{2}\bigg]
            + \frac{\kappa}{N^{2}} \bigg).
        \end{align*}
        Thus, the claim is proved.
\end{proof}

\subsection{Proof of Theorem \ref{thm:norm_MC,RQMC} for the MLMC-based Method}
\label{thm4:proof}
\begin{proof}
By taking the same result from the above proof of Theorem \ref{thm:norm_MC,RQMC}, we obtain
\begin{align*}
    \mathcal{F}(\lambda_{t+1}) - \mathcal{F}(\lambda_{t})
            &\leq - \alpha_{t} \nabla_{\lambda_{t}} \mathcal{F}(\lambda_{t})^{\top} \hat{g}_{\lambda_{t}}(\epsilon_{1:N_{t}})
            + \frac{K_{2}}{2} ||- \alpha_{t} \hat{g}_{\lambda_{t}}(\epsilon_{1:N_{t}})||_{2}^{2}.
\end{align*}
By taking the expectation with respect to $\epsilon_{1:N} \sim p(\epsilon)$, we obtain
        \begin{align*}
            \mathbb{E}_{p(\epsilon_{1:N_{t}})}&[\mathcal{F}(\lambda_{t+1}) - \mathcal{F}(\lambda_{t})] \notag \\
            &\leq - \alpha_{t} \nabla_{\lambda_{t}} \mathcal{F}(\lambda_{t})^{\top} \mathbb{E}_{p(\epsilon_{1:N_{t}})}[\hat{g}_{\lambda_{t}}(\epsilon_{1:N_{t}})]
            + \frac{K_{2}}{2} \mathbb{E}_{p(\epsilon_{1:N})}\bigg[ || -\alpha_{t}\hat{g}_{\lambda_{t}}(\epsilon_{1:N_{t}})||_{2}^{2} \bigg].
        \end{align*}
Using the fact that $\mathbb{E}_{p(\epsilon_{1:N})}[ || -\alpha_{t}\hat{g}_{\lambda_{t}}(\epsilon_{1:N})||_{2}^{2}] = 
        \mathbb{V}[-\alpha_{t}\hat{g}_{\lambda_{t}}(\epsilon_{1:N})] + ||\mathbb{E}_{p(\epsilon_{1:N})} [-\alpha_{t}\hat{g}_{\lambda_{t}}(\epsilon_{1:N})]||_{2}^{2}$ and $\mathbb{E}_{p(\epsilon_{1:N_{t}})}[\hat{g}_{\lambda_{t}}(\epsilon_{1:N_{t}})] = \nabla_{\lambda_{t}}\mathcal{F}(\lambda_{t})$, we obtain
        \begin{align}
        \label{thm3:eq1}
            \mathbb{E}_{p(\epsilon_{1:N_{t}})}&[\mathcal{F}(\lambda_{t+1}) - \mathcal{F}(\lambda_{t})] \notag \\
            &\leq - \alpha_{t} ||\nabla_{\lambda_{t}} \mathcal{F}(\lambda_{t})||_{2}^{2} + \frac{K_{2}}{2}
            \bigg(\mathbb{V}[-\alpha_{t}\hat{g}_{\lambda_{t}}(\epsilon_{1:N_{t}})] + \alpha_{t}^{2}||\mathbb{E}_{p(\epsilon_{1:N_{t}})} [\hat{g}_{\lambda_{t}}(\epsilon_{1:N_{t}})]||_{2}^{2} \bigg).
        \end{align}
According to Lemma \ref{lem:MLRG_update}, we obtain $-\alpha_{t}\hat{g}_{\lambda_{t}}(\epsilon_{1:N_{t}}) = \frac{\eta_{t}}{\eta_{t-1}}(\lambda_{t} - \lambda_{t-1}) - \alpha_{t}\hat{g}'_{\lambda_{t}}(\epsilon_{1:N_{t}})$,
where $\hat{g}'_{\lambda_{t}}(\epsilon_{1:N_{t}}) = N_{t}^{-1} \sum_{n=1}^{N_{t}} [g_{\lambda_{t}}(\epsilon_{(n,t)}) - g_{\lambda_{t-1}}(\epsilon_{(n,t)})]$.
By using this equation and the result of Lemma~\ref{lem:var_order}, we obtain
\begin{align*}
    \mathbb{V}[-\alpha_{t}\hat{g}_{\lambda_{t}}(\epsilon_{1:N_{t}})] = \mathbb{V}\bigg[\frac{\eta_{t}}{\eta_{t-1}}(\lambda_{t} - \lambda_{t-1}) - \alpha_{t}\hat{g}'_{\lambda_{t}}(\epsilon_{1:N_{t}}) \bigg]
    = \alpha_{t}^{2} \mathbb{V}[\hat{g}'_{\lambda_{t}}(\epsilon_{1:N_{t}})]
    \leq \alpha_{t}^{2} \cdot \kappa \eta_{t-1}^{2}N_{t}^{-1}.
\end{align*}
Therefore, Eq.~\ref{thm3:eq1} can be rewritten as
\begin{align*}
    \mathbb{E}_{p(\epsilon_{1:N_{t}})}&[\mathcal{F}(\lambda_{t+1}) - \mathcal{F}(\lambda_{t})] \\
            &\leq - \alpha_{t} ||\nabla_{\lambda_{t}} \mathcal{F}(\lambda_{t})||_{2}^{2} + \frac{K_{2}}{2}
            \bigg(\mathbb{V}[-\alpha_{t}\hat{g}_{\lambda_{t}}(\epsilon_{1:N_{t}})] + \alpha_{t}^{2}||\mathbb{E}_{p(\epsilon_{1:N_{t}})} [\hat{g}_{\lambda_{t}}(\epsilon_{1:N_{t}})]||_{2}^{2} \bigg) \\
            &\leq - \alpha_{t} ||\nabla_{\lambda_{t}} \mathcal{F}(\lambda_{t})||_{2}^{2} + \frac{K_{2}}{2}\bigg( \alpha_{t}^{2} \cdot \kappa \eta_{t-1}^{2}N_{t}^{-1} + \alpha_{t}^{2}||\mathbb{E}_{p(\epsilon_{1:N_{t}})} [\hat{g}_{\lambda_{t}}(\epsilon_{1:N_{t}})]||_{2}^{2} \bigg) \\
            &= - \alpha_{t} ||\nabla_{\lambda_{t}} \mathcal{F}(\lambda_{t})||_{2}^{2} + \frac{\alpha_{t}^{2}K_{2}}{2}\bigg( \kappa \eta_{t-1}^{2}N_{t}^{-1} + ||\mathbb{E}_{p(\epsilon_{1:N_{t}})} [\hat{g}_{\lambda_{t}}(\epsilon_{1:N_{t}})]||_{2}^{2} \bigg).
\end{align*}
Using the fact that $\mathbb{E}_{p(\epsilon_{1:N})} [\hat{g}_{\lambda_{t}}(\epsilon_{1:N})]
        = \nabla_{\lambda_{t}} \mathcal{F}(\lambda_{t})$, we can rewrite the above equation as
\begin{align*}
    \mathbb{E}_{p(\epsilon_{1:N_{t}})}[\mathcal{F}(\lambda_{t+1}) - \mathcal{F}(\lambda_{t})] &\leq \frac{\kappa K_{2} \alpha_{t}^{2}}{2N_{t}} \cdot \eta_{t-1}^{2} + \bigg( \frac{\alpha_{t}^{2}K_{2}}{2} - \alpha_{t}  \bigg)||\nabla_{\lambda_{t}} \mathcal{F}(\lambda_{t})||_{2}^{2}.
\end{align*}
By summing for $t=1,2,\dots,T$ and taking the total expectation, we obtain
\begin{align*}
    \mathbb{E}[\mathcal{F}(\lambda_{T}) - \mathcal{F}(\lambda_{1})] &\leq \frac{\kappa K_{2}}{2} \sum_{t=1}^{T} \frac{\alpha_{t}^{2}}{N_{t}} \eta_{t-1}^{2} + \sum_{t=1}^{T} \bigg( \frac{\alpha_{t}^{2}K_{2}}{2} - \alpha_{t}  \bigg) \mathbb{E}\bigg[||\nabla_{\lambda_{t}} \mathcal{F}(\lambda_{t})||_{2}^{2}\bigg].
\end{align*}
From the fact that $\mathcal{F}(\lambda^{*}) - \mathcal{F}(\lambda_{1}) \leq \mathbb{E}[\mathcal{F}(\lambda_{T}) - \mathcal{F}(\lambda_{1})]$, we obtain the following inequality by dividing the inequality by $A_{T} = \sum_{t=1}^{T} \alpha_{t}$:
\begin{align*}
    \frac{1}{A_{T}} [\mathcal{F}(\lambda^{*}) - \mathcal{F}(\lambda_{1})] \leq \frac{\kappa K_{2}}{2A_{T}} \sum_{t=1}^{T} \frac{\alpha_{t}^{2}}{N_{t}} \eta_{t-1}^{2} + \frac{1}{A_{T}} \sum_{t=1}^{T} \bigg( \frac{\alpha_{t}^{2}K_{2}}{2} - \alpha_{t}  \bigg) \mathbb{E}\bigg[||\nabla_{\lambda_{t}} \mathcal{F}(\lambda_{t})||_{2}^{2}\bigg].
\end{align*}
Therefore, we can obtain
\begin{align*}
    \frac{1}{A_{T}} &\sum_{t=1}^{T} \alpha_{t} ||\nabla_{\lambda_{t}} \mathcal{F}(\lambda_{t})||_{2}^{2} \\
    &\leq \frac{1}{A_{T}} [\mathcal{F}(\lambda_{1}) - \mathcal{F}(\lambda^{*})] + \frac{K_{2}}{2A_{T}} \sum_{t=1}^{T} \alpha_{t}^{2} \mathbb{E}\bigg[||\nabla_{\lambda_{t}} \mathcal{F}(\lambda_{t})||_{2}^{2}\bigg] + \frac{\kappa K_{2}}{2A_{T}} \sum_{t=1}^{T} \frac{\alpha_{t}^{2}}{N_{t}} \eta_{t-1}^{2} \\
    &= \frac{1}{A_{T}} [\mathcal{F}(\lambda_{1}) - \mathcal{F}(\lambda^{*})] + \frac{\alpha_{0}^{2} K_{2}}{2A_{T}} \sum_{t=1}^{T} \eta_{t}^{2} \bigg( \mathbb{E}\bigg[||\nabla_{\lambda_{t}} \mathcal{F}(\lambda_{t})||_{2}^{2}\bigg] + \frac{\kappa}{N_{t}} \eta_{t-1}^{2} \bigg),
\end{align*}
and the claim is proved.
\end{proof}

\subsection{Proof of Theorem \ref{prop:SNR}}
\label{prop2:proof}
\begin{proof}
Firstly, we focus on the MC- and RQMC-based methods.
Because of the definition of SNR and $\mathbb{E}_{p(\epsilon)}[\hat{g}_{\lambda}(\epsilon_{1:N})] = \nabla_{\lambda_{t}}\mathcal{F}(\lambda_{t})$,
we obtain
\begin{align*}
    \mathrm{SNR}(\lambda) &= \frac{\|\mathbb{E}_{p(\epsilon_{1:N})}[\hat{g}_{\lambda}(\epsilon_{1:N})] \|_{2}^{2}}{\sqrt{\mathbb{V}[\hat{g}_{\lambda}(\epsilon_{1:N})]}}
    = \frac{\| \nabla_{\lambda}\mathcal{F}(\lambda) \|_{2}^{2}}{\sqrt{\mathbb{V}[\hat{g}_{\lambda}(\epsilon_{1:N})]}}.
\end{align*}
Suppose that Assumptions \ref{asm:gaussian}--\ref{asm:robbins_monro} hold, the expectation and variance of the gradient estimator are nonzero, and the variances of $\hat{g}_{\lambda}(\epsilon_{1:N})$ are also nonzero.
Then, we have the upper bounds of gradient variance for the MC- and RQMC-based methods as, for all $\lambda_{t}$, $\mathbb{V}[\hat{g}_{\lambda_{t}}] \leq \kappa N^{-1}$ and $\mathbb{V}[\hat{g}_{\lambda_{t}}] \leq \kappa N^{-2}$, respectively, where $\kappa$ is a positive constant.
By using the order of gradient variance in the above, we obtain the following lower bounds:
\begin{align*}
    \mathrm{SNR}(\lambda_{t}) = \frac{\| \nabla_{\lambda_{t}}\mathcal{F}(\lambda_{t}) \|_{2}^{2}}{\sqrt{\mathbb{V}[\hat{g}_{\lambda_{t}}(\epsilon_{1:N})]}}
    \geq 
    \begin{cases}
    \frac{\| \nabla_{\lambda_{t}}\mathcal{F}(\lambda_{t}) \|_{2}^{2}}{\sqrt{\kappa N^{-1}}}, \\
    \frac{\| \nabla_{\lambda_{t}}\mathcal{F}(\lambda_{t}) \|_{2}^{2}}{\sqrt{\kappa N^{-2}}},
    \end{cases} 
    = \begin{cases}
    \frac{\| \nabla_{\lambda_{t}}\mathcal{F}(\lambda_{t}) \|_{2}^{2}}{\sqrt{\kappa}} \cdot \sqrt{N} & (\mathrm{MC}), \\
    \frac{\| \nabla_{\lambda_{t}}\mathcal{F}(\lambda_{t}) \|_{2}^{2}}{\sqrt{\kappa}} \cdot N & (\mathrm{RQMC}).
    \end{cases}
\end{align*}

Secondly, we show the SNR bound of our method.
For our method, the SNR can be expressed as
\begin{align*}
    \mathrm{SNR}(\lambda_{t}) = \frac{\|\mathbb{E}_{p(\epsilon_{1:N})}[\hat{g}_{\lambda}(\epsilon_{1:N})] \|_{2}^{2}}{\sqrt{\mathbb{V}[\hat{g}_{\lambda}(\epsilon_{1:N})]}}
    = \frac{\|\nabla_{\lambda_{t}}\mathcal{F}(\lambda_{t}) \|_{2}^{2}}{\sqrt{\alpha_{t}^{-2}\mathbb{V}[\alpha_{t}\hat{g}_{\lambda}(\epsilon_{1:N})]}}.
\end{align*}
According to the proof of Theorem \ref{thm:norm_MC,RQMC} for the MLMC-based method,
\begin{align*}
\mathbb{V}[\alpha_{t}\hat{g}_{\lambda}(\epsilon_{1:N})] = \mathbb{V}[-\alpha_{t}\hat{g}_{\lambda}(\epsilon_{1:N})] 
\leq \alpha_{t}^{2} \cdot \kappa \eta_{t-1}^{2}N_{t}^{-1}.
\end{align*}
Therefore, we obtain
\begin{align*}
    \mathrm{SNR}(\lambda_{t})
    \geq \frac{\|\nabla_{\lambda_{t}}\mathcal{F}(\lambda_{t}) \|_{2}^{2}}{\sqrt{\alpha_{t}^{-2}\alpha_{t}^{2} \cdot \kappa \eta_{t-1}^{2}N_{t}^{-1}}}
    = \frac{\|\nabla_{\lambda_{t}}\mathcal{F}(\lambda_{t})  \|_{2}^{2}}{\sqrt{\kappa \eta_{t-1}^{2}N_{t}^{-1}}} 
    = \frac{\|\nabla_{\lambda_{t}}\mathcal{F}(\lambda_{t})  \|_{2}^{2}}{\sqrt{\kappa}} \cdot \frac{\sqrt{N_{t}}}{\eta_{t-1}}.
\end{align*}
Thus, the claim is proved.
\end{proof}
\section{Unbounded Case in Assumption \ref{asm:lipshitz2}}
\label{app.non-bounded}
Assumption \ref{asm:lipshitz2} does not always hold because $\mathcal{T}(\epsilon;\lambda)$ itself is not bounded.
To overcome this problem, the proximal operator is useful.
The proximal operator is an operator associated with a closed convex function $f$ from a Hilbert space $\mathcal{X}$ to $[-\infty, \infty]$, and it is defined as
\begin{align*}
    \mathrm{prox}_{f}(x) = \argmin_{u\in\mathcal{X}} \bigg( f(u) + \frac{1}{2} \| u - x \|_{2}^{2} \bigg).
\end{align*}
In optimization, the proximal operator has several useful properties such as its firm nonexpansiveness as follow:
\begin{align*}
    \| \mathrm{prox}_{f}(x) - \mathrm{prox}_{f}(y)\|_{2}^{2} \leq (\mathrm{prox}_{f}(x) - \mathrm{prox}_{f}(y))^{\top} (x-y),
\end{align*}
where $\forall x,y \in \mathcal{X}$.

We will apply this to the case where $\mathcal{T}(\epsilon;\lambda)$ becomes too large to satisfy Assumption~\ref{asm:lipshitz2} in our method.
We set $f$ as the following indicator function for some set $S$:
\begin{align*}
    \mathrm{I}_{S}(\mathcal{T}(\epsilon;\lambda)) = 
    \begin{cases}
        0 & (\mathcal{T}(\epsilon;\lambda) \in S), \\
        +\infty & (\mathrm{otherwise}),
    \end{cases}
\end{align*}
where $\mathrm{I}_{S}(\mathcal{T}(\epsilon;\lambda))$ is closed and convex if $S$ is a closed convex set.
From this setting, the proximal operator $f=\mathrm{I}_{S}$ is the Euclidian projection $P(\cdot)$ on $S$:
\begin{align*}
    \mathrm{prox}_{\mathrm{I}_{S}}(\mathcal{T}(\epsilon;\lambda)) = \argmin_{u \in S} \| u - \mathcal{T}(\epsilon;\lambda) \|_{2}^{2} 
    = P_{S}(\mathcal{T}(\epsilon;\lambda)).
\end{align*}
From the above and firm nonexpansiveness of the proximal operator, if $\mathcal{T}^{+}(\epsilon;\lambda) = \mathrm{prox}_{\mathrm{I}_{S}}(\mathcal{T}(\epsilon;\lambda))$ and $\mathcal{T}^{+}(\epsilon;\bar{\lambda}) = \mathrm{prox}_{\mathrm{I}_{S}}(\mathcal{T}(\epsilon;\bar{\lambda}))$, then
\begin{align*}
    \|\mathcal{T}^{+}(\epsilon;\lambda) - \mathcal{T}^{+}(\epsilon;\bar{\lambda}) \|_{2}^{2}
    \leq (\mathcal{T}^{+}(\epsilon;\lambda) - \mathcal{T}^{+}(\epsilon;\bar{\lambda}))^{\top}(\mathcal{T}(\epsilon;\lambda) - \mathcal{T}(\epsilon;\bar{\lambda}))
\end{align*}
is fulfilled.
This implies that
\begin{align*}
    \|\mathcal{T}^{+}(\epsilon;\lambda) - \mathcal{T}^{+}(\epsilon;\bar{\lambda}) \|_{2}^{2} \leq \|\mathcal{T}(\epsilon;\lambda) - \mathcal{T}^{+}(\epsilon;\bar{\lambda}) \|_{2}^{2}
\end{align*}
from the Cauchy-–Schwarz inequality.
It means that $\mathcal{T}^{+}(\epsilon;\lambda)$ is 1-Lipschitz continuous and therefore bounded.

By using the proximal operator, we can obtain the samples that do not violate Assumption~\ref{asm:lipshitz2}.
Therefore, all the theorems and lemmas in this paper hold in a unbounded case of $\mathcal{T}(\epsilon;\lambda)$.
The MLMCVI algorithm in a unbounded case is shown in Algorithm \ref{alg2}.
\begin{algorithm}[t]                      
                \caption{Multilevel Monte Carlo Variational Inference in Unbounded Case of $\mathcal{T}(\epsilon;\lambda)$}
                \label{alg2}
                \begin{algorithmic}[1]
                        \REQUIRE {Data ${\bf x}$, random variable $\epsilon \sim p(\epsilon)$, transform ${\bf z} = \mathcal{T}(\epsilon;\lambda)$, model $p({\bf x},{\bf z})$, 
                        variational family $q({\bf z}|\lambda)$}
                        \ENSURE {Variational parameter $\lambda^{*}$}
                        \STATE \textbf{Initialize:} $N_{0}$, $\lambda_{0}$, $\alpha_{0}$, the hyperparameter of $\eta$, and a convex set $S$
                        \FOR {$t = 0$ to $T$}
                        \IF {$t=0$}
                        \STATE $\epsilon_{n} \sim p(\epsilon) \ (n = 1,2,\dots,N_{0})$ \ $\triangleleft$ \ \textrm{sampling} $\epsilon$
                        \STATE $\mathcal{T}^{+}(\epsilon_{n};\lambda_{0}) = \mathrm{prox}_{\mathrm{I}_{S}}(\mathcal{T}(\epsilon_{n};\lambda_{0}))$ \ $\triangleleft$ \ \textrm{check $\mathcal{T}(\epsilon;\lambda_{0})$ value}
                        \STATE $\hat{g}_{\lambda_{0}}(\epsilon_{1:N_{0}}) = N_{0}^{-1} \sum_{n=1}^{N_{0}} g_{\lambda_{0}}(\epsilon_{n})$ \ $\triangleleft$ \ \textrm{calc. RG estimator}
                        \STATE $\lambda_{1} = \lambda_{0} - \alpha_{0} \hat{g}_{\lambda_{0}}(\epsilon_{1:N_{0}})$ \ $\triangleleft$ \ \textrm{grad-update}
                        \ELSE
                        \STATE \textbf{estimate} $N_{t}$ using $N_{t} =  \lceil{\eta_{t-1} N_{0} \rceil}$
                        \STATE \textbf{sampling} one $\epsilon$ for sample size estimation
                        \STATE $\epsilon_{n} \sim p(\epsilon) \ (n = 1,2,\dots,N_{t})$ \ $\triangleleft$ \ \textrm{sampling} $\epsilon$
                        \STATE $\mathcal{T}^{+}(\epsilon_{n};\lambda_{t}) = \mathrm{prox}_{\mathrm{I}_{S}}(\mathcal{T}(\epsilon_{n};\lambda_{t}))$ \ $\triangleleft$ \ \textrm{check $\mathcal{T}(\epsilon;\lambda_{t})$ value}
                        \STATE $\hat{g}^{'}_{\lambda_{t}}(\epsilon_{1:N_{t}}) = N_{t}^{-1} \sum_{n=1}^{N_{t}} [g_{\lambda_{t}}(\epsilon_{(n,t)}) - g_{\lambda_{t-1}}(\epsilon_{(n,t)}) ]$
                        \ $\triangleleft$ \ \textrm{calc. Multilevel term}
                        \STATE $\lambda_{t+1} = \lambda_{t} + \frac{\eta_{t}}{\eta_{t-1}}(\lambda_{t} - \lambda_{t-1}) - \alpha_{t} \hat{g}^{'}_{\lambda_{t}}(\epsilon_{1:N_{t}})$ \ $\triangleleft$ \ \textrm{grad-update}
                        \IF {$\lambda_{t+1}$ \textrm{has converged to} $\lambda^{*}$}
                        \STATE \textbf{break}
                        \ENDIF
                        \ENDIF
                        \ENDFOR
                        \RETURN $\lambda^{*}$
                \end{algorithmic}
        \end{algorithm}

One possible problem with artificially truncating the value of $\mathcal{T}(\epsilon;\lambda)$ is that the gradient estimator can be biased because multiple values of the samples $\epsilon$ may be mapped to the same bounded value of $\mathcal{T}(\epsilon;\lambda)$.
However, in practice, we can reduce this bias by taking a sufficiently large closed convex set $S$ so that we can take $u \in S$ more flexible.
Furthermore, the truncation through the proximal operator can be seen as the ``gradient clipping'' technique that has been empirically successful, especially in deep learning context~\citep{mikolov12,pascanu13,kanai17,belghazi18a}.
From these empirical successes, the gradient truncation through a proximal operator is one realistic way to ensure that Assumption~\ref{asm:lipshitz2} is satisfied.
A summary of the gradient clipping can be seen in Section 10.11.1 of the paper by \cite{Goodfellow16}.

Theoretical analyses of the properties of gradient clipping, such as analysis of bias due to truncation by the proximal operator, is beyond the scope of our paper. 
However, such theoretical analyses are important for improving the performance of optimization through gradient clipping.
Research on why gradient clipping works well has recently been conducted, e.g., \citet{Zhang20}, and such research on the theoretical exploration of gradient clipping is beginning to attract attention.
Thus, it may be very interesting for future research to discuss how the bias due to gradient clipping, such as truncation, affects stochastic optimization in the MCVI context.

\section{Learning Rate Scheduler and Sample Size Estimation}
\label{sec:another_samplesize}
From Algorithm~\ref{alg3}, we can estimate the number of samples $N_{t}$ as $N_{t+1} = \lceil{ \eta_{t} N_{1} \rceil}$ and $N_{0} = N_{1}$ from Theorem~\ref{thm:alt_optimal_sample} and Lemma~\ref{cor:optimal}.
Therefore, its estimation scheme varies with the learning rate scheduler function $\eta_{t-1}$.
There are three major learning rate schedulers: time-based decay, step-based decay, and exponential decay defined as follows.
\begin{definition}[Time-based Decay Function]
Time-based decay function $\eta_{t}$ is defined as $\eta_{t} = \frac{1}{1+\beta t}$,
where $\beta$ is the parameter of the degree of decay.
\end{definition}

\begin{definition}[Step-based Decay Function]
Step-based decay function $\eta_{t}$ is defined as $\eta_{t} = \beta^{\lfloor{ \frac{t}{r} \rfloor}}$,
where $\beta$ is the parameter of the degree of decay, and $r$ is the drop-rate parameter.
Here, we denote the function $\lfloor x \rfloor$ as $\lfloor x \rfloor = \max \{k \in \mathbb{Z}| k \leq x \}$.
\end{definition}

\begin{definition}[Exponential Decay Function]
Exponential decay function $\eta_{t}$ is defined as $\eta_{t} = \exp(-\beta t)$,
where $\beta$ is the parameter of the degree of decay.
\end{definition}
In these functions, $N_{t}$ for $t \geq 1$ is estimated as follows:
\begin{itemize}
    \item Time-based decay: $N_{t} = \lceil{ \frac{1}{1+\beta (t-1)} N_{0} \rceil}$,
    \item Step-based decay: $N_{t} = \lceil{ \beta^{\lfloor{\frac{t-1}{r} \rfloor}} N_{0} \rceil}$,
    \item Exponential decay: $N_{t} = \lceil{ \exp({-\beta(t-1) }) N_{0} \rceil}$,
\end{itemize}
where $\beta$ and $r$ are the decay and drop-rate parameters, respectively.

\section{Memory Space Cost and Time Cost for the MLRG Estimator}
\label{app:memory_cost}
In this section, we discuss how our method affects the memory cost {of} gradient estimation.
Since the MC- and RQMC-based methods estimate the ELBO {or the variational free energy} using the current parameters and evaluate its stochastic gradient, the {required} cost of memory space is $\mathcal{O}(d)$ per iteration, where $d$ is the dimension of the parameter.
Furthermore, the time cost can be seen as $\mathcal{O}(N \times d)$, where $N$ is the sample size for gradient estimation.

{In contrast}, in our method, we should {keep} the old parameter and estimate the old gradient per iteration; hence, we can see that the {required} cost of memory space is $\mathcal{O}(2d)$.
In addition, the time cost of our method can be seen as $\mathcal{O}(2 \times N_{t} \times d)$.
We summarize the {memory space} cost and the time cost of each method in Table~\ref{tab:cost}.

\begin{figure}[ht]
    \centering
    \includegraphics[scale=0.6]{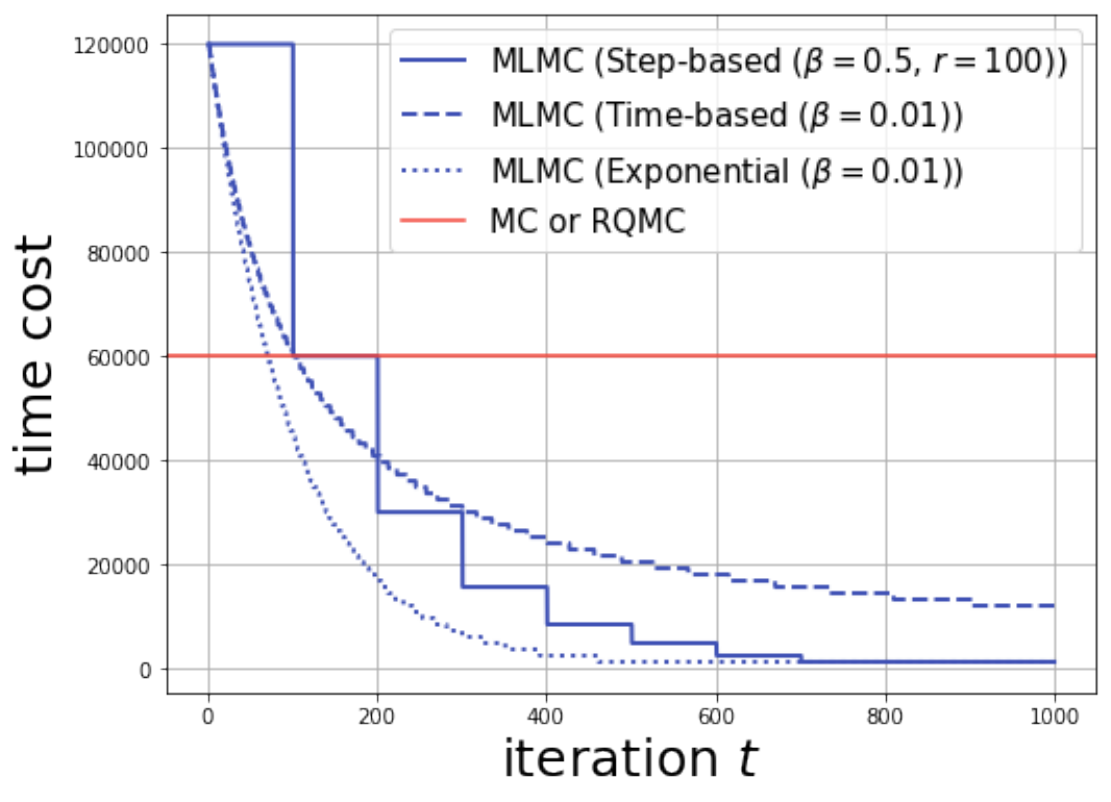}
    \caption{Time cost vs. iteration plot. }
    \label{fig:time_cost}
\end{figure}

{It may seem that the time cost of our method is too large compared with that of the baseline methods.}
However, as shown in Figure~\ref{fig:time_cost}, the time cost of our method becomes smaller than that of the baseline methods as the optimization proceeds because our method decreases the sample size for gradient estimation using a learning rate scheduler $\eta$.

{Therefore, the hyperparameter optimization of the learning rate scheduler is important in our method.}
If the sample size is reduced too {fast}, the time cost will be decreased quickly; however, the inference may be affected.
On the other hand, if the sample size is reduced too slowly, the estimation {can} be stable, but the time cost will be large through optimization.
{Unfortunately, theoretically deriving the optimal hyperparameters is an open problem in stochastic optimization.}
{One practical way to optimize a hyperparameter of the learning rate scheduler is using a hyperparameter optimization tool such as {\sf optuna}~\citep{optuna_2019}.}
\begin{table}[t]
\centering
\begin{tabular}{ccc}
\hline
\hline
                      & \textbf{MC or RQMC}       & \textbf{MLMC}                          \\ \hline
\textbf{Time cost}    & $\mathcal{O}(N \times d)$ & $\mathcal{O}(2 \times N_{t} \times d)$ \\
\textbf{Memory space} & $\mathcal{O}(d)$          & $\mathcal{O}(2 \times d)$                     
\end{tabular}
\caption{Summary of time cost and cost of memory space.~\label{tab:cost}}
\end{table}
\section{Details of Models in Experiments}
In this section, we show the details of the model generative process and hyperparameter settings used in our experiments.

\subsection{Hierarchical Linear Regression}
\label{hbr_setting}
We applied hierarchical linear regression to toy data generated from the same generating process of this model.
Here, we set a Gaussian hyperprior on $\mu'$
and lognormal hyperpriors on the variance of intercepts $\sigma'$ and the noise $\epsilon$.

The generative process of this model is as follows.
\begin{align*}
    &\mu' \sim \mathcal{N}(0, 10^{2}), \ \ &\textrm{weight hyperprior} \\
    &\sigma' \sim \textrm{LogNormal}(0.5), \ \ &\textrm{weight hyperprior} \\
    &\epsilon \sim \textrm{LogNormal}(0.5), \ \ &\textrm{noise} \\
    &{\bf b}_{i} \sim \mathcal{N}(\mu',\sigma'), \ \ &\textrm{weights} \\
    &y_{i} \sim \mathcal{N}({\bf x}_{i}^{\top} {\bf b}_{i}, \epsilon). \ \ &\textrm{output distribution}
\end{align*}

We set $I=100$ and $k=10$, where $k$ denotes the dimension of the data $x_{i}$ and $I$ is the number of observations.
In this setting, the dimension of the entire parameter space is $d = I \times k + k + 2 = 1012$, and this model is approximated by a variational diagonal Gaussian distribution.

We optimized the {variational free energy} of the MC- and RQMC-based methods by using the Adam optimizer~\citep{kingma14b} and that of the MLMC-based method by using the SGD optimizer with a learning rate scheduler $\eta$ with $100$ initial MC or RQMC samples.
We compared the empirical variance and SNR of these methods by using $100$ or $10$ initial MC or RQMC samples for inference.
In the optimization step, we used $\eta$ as the step-decay function. 
The initial learning rate ($\in [10^{-3},0.5]$) and the hyperparameters of {the} learning rate scheduler, $\beta$ ($\in [0.1,1]$) and $r$ ($\in[10,500]$), were optimized through the Tree-structured Parzen Estimator (TPE) sampler in {\sf{optuna}}~\citep{optuna_2019} with $50$ trials, where $\beta$ and $r$ are the decay and drop-rate parameter, respectively.

\subsection{Bayesian Logistic Regression}
\label{blr_setting}

\paragraph{Binary Classification:}
We applied Bayesian Logistic Regression to the breast cancer dataset in the UCI Machine Learning Repository for a binary classification task.
Here, we set a standard Gaussian hyperprior on $\mu'$ and an inverse gamma hyperprior (weak information prior) on the variance of weights $\sigma'$.

The generative process of this model is as follows.
\begin{align*}
    &\sigma' \sim \textrm{Gamma}(0.5, 0.5), \ \ &\textrm{weight hyperprior} \\
    &\mu' \sim \mathcal{N}(0,1), \ \ &\textrm{weight hyperprior} \\
    &{\bf z}_{i} \sim \mathcal{N}(\mu',1/\sigma'), \ \ &\textrm{weights} \\
    &\sigma(x) = \frac{1}{1 + \exp(-x)},  \ \ &\textrm{Sigmoid function} \\
    &y_{i} \sim \textrm{Bernoulli}(\sigma({\bf x}_{i}^{\top} {\bf z}_{i})). \ \ &\textrm{output distributions}
\end{align*}

In these settings, the dimension of the entire parameter space is $d = 11$, and this model is approximated by a variational diagonal Gaussian distribution.

To optimize the {variational free energy}, we used the Adam optimizer for the MC- and RQMC-based methods and the SGD optimizer with a learning rate scheduler $\eta$ for the MLMC-based method.
We used $100$ initial MC or RQMC samples for gradient estimation.

We compared the empirical variance and SNR of these methods by using $100$ or $10$ initial MC or RQMC samples for inference.
In the optimization step, we used $\eta$ as the step-decay function. 
The initial learning rate ($\in[10^{-5},10^{-2}]$) and the hyperparameters of {the} learning rate scheduler, $\beta$ ($\in [0.1,1]$) and $r$ ($\in[10,500]$), were optimized through the Tree-structured Parzen Estimator (TPE) sampler in {\sf{optuna}}~\citep{optuna_2019} with $50$ trials, where $\beta$ and $r$ are the decay and drop-rate parameter, respectively.

\paragraph{Multilabel Classification:}
We also applied this model to the Fashion-MNIST dataset for a multilabel classification task.
We set a standard Gaussian hyperprior on $\mu'$ and an inverse gamma hyperprior (weak information prior) on the variance of weights $\sigma'$.
Thus, the generative process of this model is as follows.
\begin{align*}
    &\sigma' \sim \textrm{Gamma}(0.5, 0.5), \ \ &\textrm{weight hyperprior} \\
    &\mu' \sim \mathcal{N}(0,1), \ \ &\textrm{weight hyperprior} \\
    &{\bf z}_{i} \sim \mathcal{N}(\mu',1/\sigma'), \ \ &\textrm{weights} \\
    &\sigma(x_{i}) = \frac{\exp(x_{i})}{\sum_{j} \exp(x_{j})}, \ \ &\textrm{Softmax function} \\
    &y \sim \textrm{Categorical}(\sigma(\phi({\bf x}_{i}^{\top}{\bf z}_{i}))). \ \ &\textrm{output distributions}
\end{align*}

In these settings, the dimension of the entire parameter space is $d = 7852$, and this model is also approximated by a variational diagonal Gaussian distribution.

We optimized the {variational free energy} of the MC-, RQMC-, and MLMC-based methods by using the SGD optimizer with a learning rate scheduler $\eta$. We used $100$ initial MC or RQMC samples for gradient estimation.
In the optimization step, we used $\eta$ as the step-decay function and set the hyperparameter $\{\beta, r\}$ for sample size estimation to $\{0.5, 100 \}$.
Finally, we set the initial learning rate as $0.01$, $0.005$, or $0.001$.

\subsection{Bayesian Neural Network Regression}
\label{bnnreg_setting}
We applied a BNN regression model to the wine-quality-red dataset, which is included the wine-quality dataset in the UCI Machine Learning Repository.

The network consists of a $50$-unit hidden layer with ReLU activations.
In addition, we set a normal prior over each weight and placed an inverse Gamma hyperprior
over each weight prior, and we also set an inverse Gamma hyperprior to the observed variance.

The generative process of this model is as follows.
\begin{align*}
    &\alpha \sim \textrm{Gamma}(1., 0.1), \ \ &\textrm{weight hyperprior} \\
    &\tau \sim \textrm{Gamma}(1., 0.1), \ \ &\textrm{noise hyperprior} \\
    &w_{i} \sim \mathcal{N}(0,1/ \alpha), \ \ &\textrm{weights} \\
    &y \sim \mathcal{N}(\phi({\bf x},{\bf w}),1/ \tau). \ \ &\textrm{output distributions}
\end{align*}

In this setting, $\phi({\bf x},{\bf w})$ is a multilayer perceptron that maps input data ${\bf x}$
to output $y$ by using the set of weights $w$, and
the set of parameters is expressed as $\theta \coloneqq ({\bf w}, \alpha, \tau)$.
The model exhibits a posterior of dimension $d = 653$ and was applied to a 100-row dataset subsampled from the wine-red dataset.

We approximated the posterior of this model by using a variational diagonal Gaussian distribution, and we used the learning rate scheduler $\eta$ as the step-decay function.
The initial learning rate ($\in[10^{-5},10^{-2}]$ (MC,RQMC) or $\in[10^{-8},10^{-5}]$ (MLMC)) and the hyperparameters of {the} learning rate scheduler, $\beta$ ($\in[0.1,1]$) and $r$ ($\in[10,500]$), were optimized through the Tree-structured Parzen Estimator (TPE) sampler in {\sf{optuna}}~\citep{optuna_2019} with $50$ trials, where $\beta$ and $r$ are the decay and drop-rate parameter, respectively.

To optimize the {variational free energy}, we used the Adam optimizer for the MC- and RQMC-based methods and the SGD optimizer with a learning rate scheduler $\eta$ for the MLMC-based method.
We used $50$ initial MC or RQMC samples for gradient estimation.
We compared the empirical variance and SNR of these methods by using $50$ or $10$ initial MC or RQMC samples for inference.
\section{Additional Experimental Results for Sensitivity of Optimization for Various Hyperparameter Settings}
\label{app:additional_exp_results}
\begin{figure}[th]
    \centering
    \includegraphics[scale=0.29]{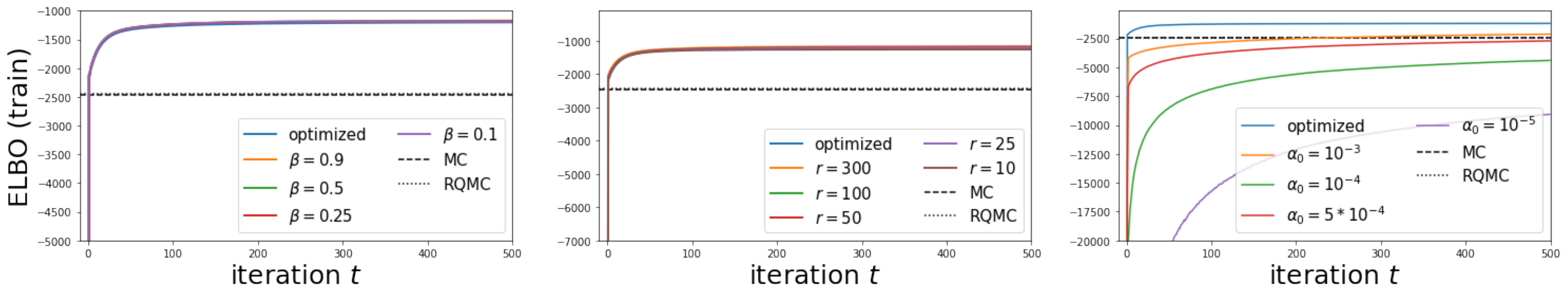}
    \caption{Experimental results for hierarchical linear regression analysis of various hyperparameter settings and initial learning rates. To confirm the optimization performance, the training ELBOs (higher is better) are lined up from the left.}
    \label{fig:hbr_all}
\end{figure}

\begin{figure}[th]
    \centering
    \includegraphics[scale=0.29]{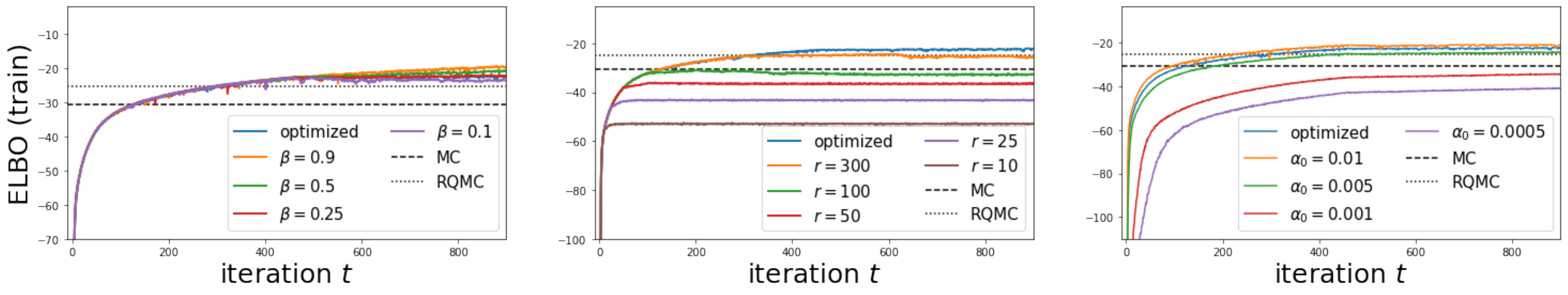}
    \caption{Experimental results for Bayesian logistic regression analysis of various hyperparameter settings and initial learning rates. To confirm the optimization performance, the training ELBOs (higher is better) are lined up from the left.}
    \label{fig:blr_all}
\end{figure}

\begin{figure}[th]
    \centering
    \includegraphics[scale=0.29]{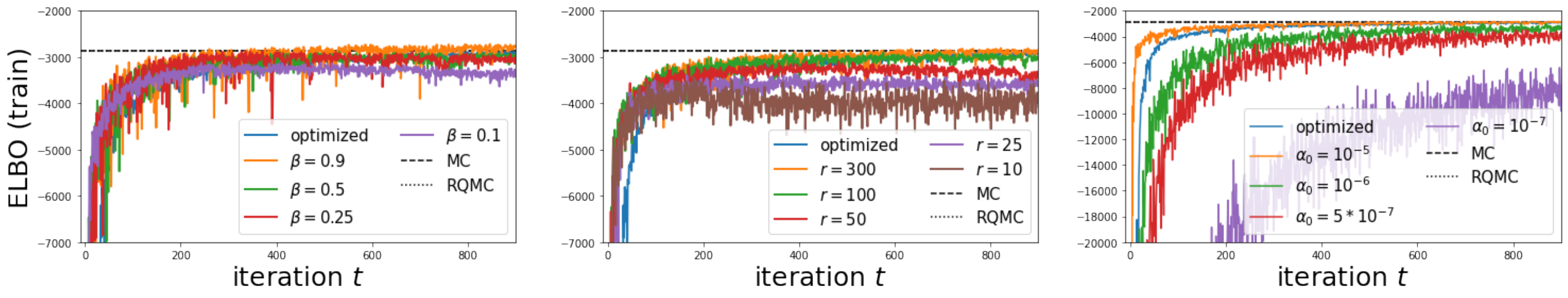}
    \caption{Experimental results for Bayesian neural network regression analysis of various hyperparameter settings and initial learning rates. To confirm the optimization performance, the training ELBOs (higher is better) are lined up from the left.}
    \label{fig:bnn_all}
\end{figure}

In this section, we conduct additional experiments to analyze our method's sensitivity for various hyperparameter settings and initial learning rates.
As we mentioned in Section~\ref{subsec:conv_anal} regarding several theoretical analyses, the optimization performance of our method depends on the learning rate scheduler function $\eta$.
To understand the characteristic of our method in more detail, we conducted benchmark experiments with various hyperparameter settings and initial learning rates to see how the performance of the optimization changes and compare it with the optimal settings in Section~\ref{subsec:exp_benchmark}.
The optimal hyperparameter settings selected by {\sf optuna} are summarized in Table~\ref{tab:hyper}.
The results are shown in Figures~\ref{fig:hbr_all}, \ref{fig:blr_all}, and \ref{fig:bnn_all}.

These experimental results show that the optimization performance of our method significantly degrades when the drop-rate $r$ and the initial learning rate are improperly set.
{This degradation reflects the property shown in Theorem~\ref{thm:norm_MC,RQMC}, which shows that the learning rate scheduler function $\eta$ affects the convergence of our method.}
Therefore, in practice, these hyperparameters should be carefully optimized according to the task and dataset in our method.
{A hyperparameter optimization library such as {\sf optuna}~\citep{optuna_2019}, which we used many times in this study, might be useful for this purpose.}

\begin{table}[t]
\centering
\scalebox{0.9}{
\begin{tabular}{c|ccc|ccc|ccc}
\hline
\hline
\textbf{}     & \multicolumn{3}{c|}{\textbf{HBR}}  & \multicolumn{3}{c|}{\textbf{BLR}}  & \multicolumn{3}{c}{\textbf{BNN}}    \\ \hline
              & $\alpha_{0}$ & $\beta$    & $r$   & $\alpha_{0}$ & $\beta$    & $r$   & $\alpha_{0}$   & $\beta$    & $r$   \\ \hline
\textbf{MC}   & $0.39893$   & -          & -     & $0.004735$   & -          & -     & $0.007780$     & -          & -     \\
\textbf{RQMC} & $0.39893$   & -          & -     & $0.007780$   & -          & -     & $0.007780$     & -          & -     \\
\textbf{MLMC} & $0.027026$   & $0.862527$ & $221$ & $0.007438$   & $0.226316$ & $458$ & $9.062263e\mathchar`-6$ & $0.819243$ & $253$
\end{tabular}
}
\caption{Selected parameters optimized by {\sf optuna}~\label{tab:hyper}}
\end{table}

\vskip 0.2in
\clearpage
\bibliography{20-653}
\end{document}